\documentclass[10pt,twocolumn,letterpaper]{article}

\usepackage{cvpr}
\usepackage{times}
\usepackage{epsfig}
\usepackage{graphicx}
\usepackage{amsmath}
\usepackage{amssymb}
\usepackage{epstopdf}
\usepackage{array}

\usepackage[breaklinks=true,bookmarks=false]{hyperref}

\cvprfinalcopy 

\ifcvprfinal\fi
\begin{document}

\title{Automated Top View Registration of Broadcast Football Videos}

\author{Rahul Anand Sharma, Bharath Bhat, Vineet Gandhi, C.V.Jawahar \\
Centre for Visual Information Technology, \\
International Institute of Information Technology, Hyderabad. INDIA. }

\maketitle

\begin{abstract}
In this paper, we propose a novel method to register football broadcast video frames on the static top view model of the playing surface. The proposed method is fully automatic in contrast to the current state of the art which requires manual initialization of point correspondences between the image and the static model. Automatic registration using existing approaches has been difficult due to the lack of sufficient point correspondences. We investigate an alternate approach exploiting the edge information from the line markings on the field. We formulate the registration problem as a nearest neighbour search over a synthetically generated dictionary of edge map and homography pairs. The synthetic dictionary generation allows us to exhaustively cover a wide variety of camera angles and positions and reduce this problem to a minimal per-frame edge map matching procedure. We show that the per-frame results can be improved in videos using an optimization framework for temporal camera stabilization. We demonstrate the efficacy of our approach by presenting extensive results on a dataset collected from matches of football World Cup 2014.

%

\end{abstract}

\section{Introduction}

\begin{figure*}[t]
\centering
\begin{tabular}[b]{c c c}
\includegraphics[width=0.29\linewidth]{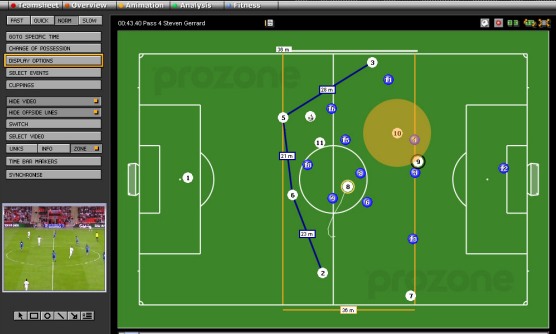} &
\includegraphics[width=0.31\linewidth]{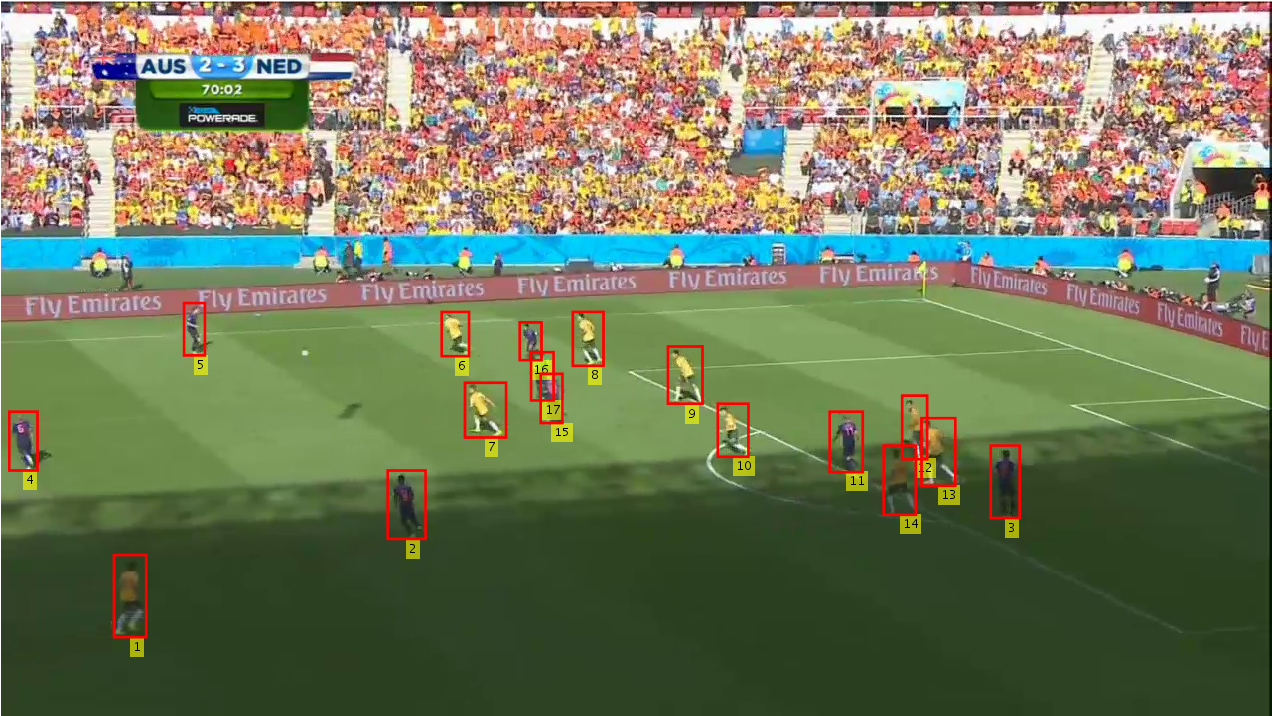} & \hspace{-1em}
\includegraphics[width=0.31\linewidth]{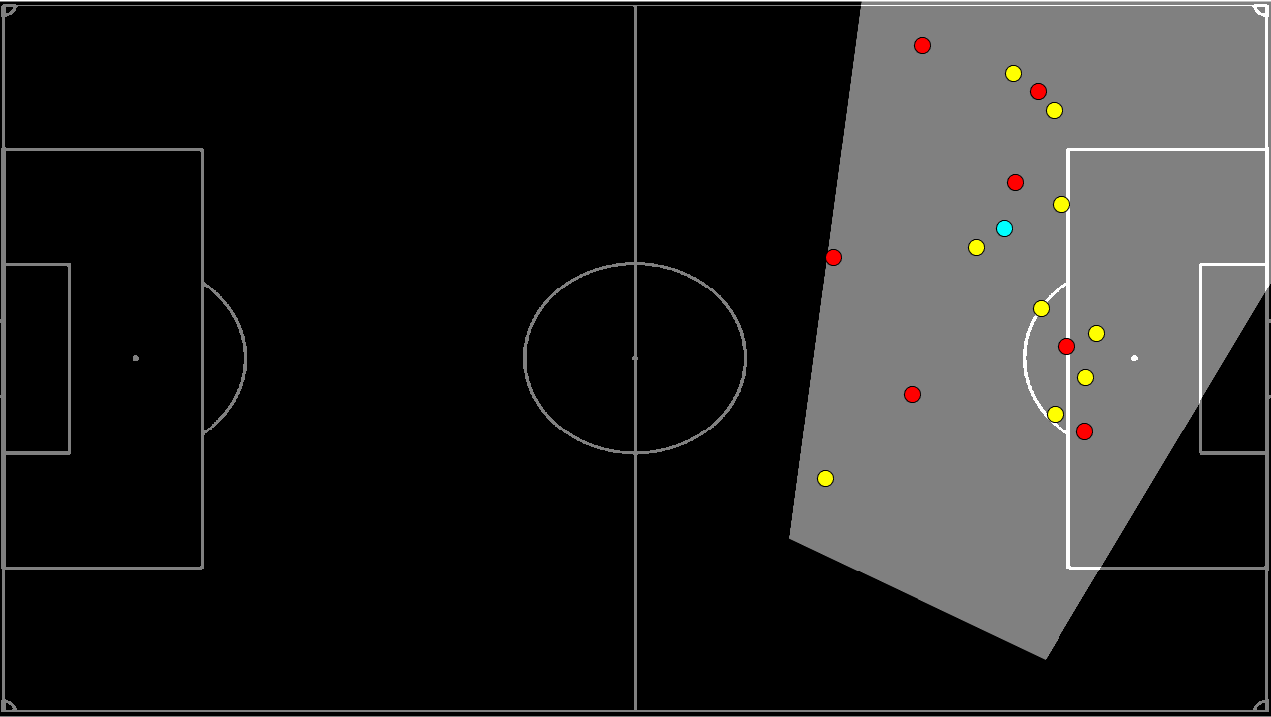} \\
(a) & \multicolumn{2}{c}{(b)} \\
\end{tabular}
\label{fig:motivation}
\caption{(a) A snapshot from prozone tracking system. (b) An example result from the proposed method, which takes as input a broadcast image and outputs its registration over the static top view model with the corresponding player positions. The yellow, red and cyan circles denote the players from different teams and referee respectively. }
\end{figure*}

Advent of tracking systems by companies like Prozone~\cite{prozone} and Tracab~\cite{tracab} has revolutionized the area of football analytics. Such systems stitch the feed from six to ten elevated cameras to record the entire football field, which is then manually labelled with player positions and identity to obtain the top view data over a static model as shown in Figure~\ref{fig:motivation}. Majority of recent research efforts~\cite{gudmundsson2016spatio,lucey2014quality,lucey2013representing,bojinovpressing} and commercial systems for football analytics have been based on such top view data (according to prozone website, more than 350 professional clubs now use their system). There are three major issues with such commercial tracking systems and associated data. First, it is highly labour and time intensive to collect such a data. Second, it is not freely available and has a large price associated with it. Third, such a data can not be obtained for analyzing matches where the customized camera installations were not used. It is also difficult for most research groups to collect their own data due to the challenges of installing and maintaining such systems and the need of specific collaborations with the clubs/stadiums.

All the above problems can be addressed, if we can obtain such data using the readily available broadcast videos. However, this is a non trivial task since the available broadcast videos are already edited and only show the match from a particular viewpoint/angle at a given time. Hence, obtaining the top view data first requires the registration of the given viewpoint with the static model of the playing surface. This registration problem is challenging because of the movement of players and the camera; zoom variations; textureless field; symmetries and highly similar regions etc. Due to these reasons, this problem has interested several computer vision researchers ~\cite{okuma2004automatic,hess2007improved,lu2013learning}, however most of the existing solutions are based on computation of point correspondences or/and require some form of manual initialization. Not just that the manual initialization for each video sequence is an impractical task (as shot changes occur quite frequently), such approaches are also not applicable in the presented scenario due to absence of good point correspondences (the football playing surface is almost textureless in contrast to the cases like American football~\cite{hess2007improved}).


Motivated by the above reasons, we take an alternate approach based on edge based features and formulate the problem as a nearest neighbour search to the closest edge map in a precomputed dictionary with known projective transforms. Since, manual labelling of a sufficiently large dictionary of edge maps with known correspondences is an extremely difficult and tedious task, we employ a semi supervised approach, where a large `camera-view edge maps to projective transform pairs' are simulated from a small set of manually annotated examples (the process is illustrated in Figure~\ref{fig:synthetic_dataset_generation}). The simulated dictionary generation allows us to cover edge maps corresponding to various degrees of movement of the camera from different viewpoints (which is an infeasible task manually). More importantly, this idea reduces the accurate homography estimation problem to a minimal dictionary search using the edge based features computed over the query image.  The tracking data can then be simply obtained by projecting the player detections performed over broadcast video frames, using the same projective transform. An example of our approach over a frame from Australia vs Netherlands world cup match is illustrated in Figure~\ref{fig:motivation}.

Since the camera follows most of the relevant events happening in the game, it can be fairly assumed that the partial tracking data (only considering the players visible in the current camera view) obtained using the proposed approach is applicable to most of the work on football play style analytics~\cite{gudmundsson2016spatio}. Furthermore, the knowledge of camera position and movement can work as an additional cue for applications like summarization and event detection (goals. corners etc.), as the camera movement and editing is highly correlated with the events happening in the game. It is also useful for content retrieval applications, for instance it can allow queries like ``give me all the counter attack shots" or ``give me all the events occurring on the top left corner" etc. The proposed approach can also be beneficial in several other interesting research topics like motion fields for predicting the evolution of the game~\cite{kim2010motion}, social saliency for optimized camera selection~\cite{soo2015social} or automated commentary generation~\cite{andre2000three}. \\

More formally this work makes following contributions:
\begin{enumerate}
\item We propose a novel framework to obtain the registration of football broadcast videos with a static model. We demonstrate that the proposed nearest neighbour search based approach makes it possible to robustly compute the homography in challenging cases, where even manually labelling the minimum four point based correspondences is difficult. 
\item We thoroughly compare three different approaches based on HOG features, chamfer matching and convolution neural net (CNN) based features to exploit the suitable edge information from the playing field. 
\item We propose a semi-supervised approach to synthetically generate a dictionary of `camera-view to projective transform pairs' and present a novel dataset with over a hundred thousand pairs. 
\item  We propose a mechanism to further enhance the results on video sequences using a Markov Random Field (MRF) optimization and a convex optimization framework for removing camera jitter .
\item We present extensive qualitative and quantitative results on a simulated and a real test dataset, to demonstrate the effectiveness of the proposed approach.
\end{enumerate}
 
The proceeding section briefly explains the related work. The semi-supervised dictionary learning approach is described in Section~\ref{sec:dictionary_generation}, followed by the explanation of the proposed matching algorithms. Section~\ref{sec:smoothing_and_stabilization} covers the optimization techniques followed by the experimental results and concluding discussion.   
\section{Related work}
\begin{figure*}[t!]
\centering
\includegraphics[width=0.9\linewidth]{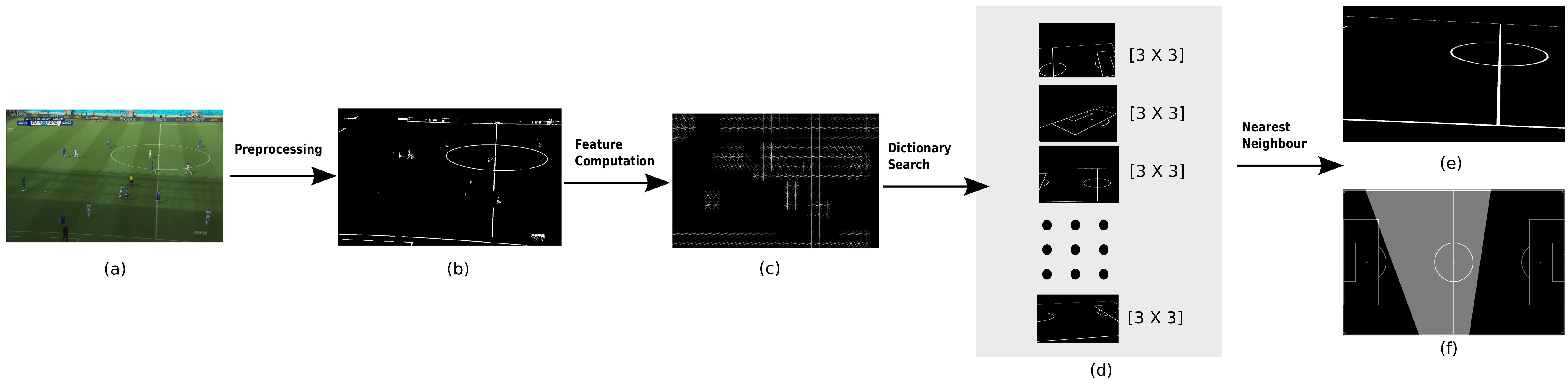}
\caption{Overview of the proposed approach. The input to the system is a broadcast image (a) and the output is the registration over the static model (f). The image (e) shows the corresponding nearest neighbour edge map from the synthetic dictionary.}
\label{fig:Overview}
\end{figure*}
Top view data for sports analytics has been extensively used in previous works. Bialkowski et al.~\cite{bialkowski2013recognising} uses 8 fixed high-definition (HD) cameras to detect the players in field hockey matches. They demonstrated that event recognition (goal, penalty corner etc.) can be performed robustly even with noisy player tracks. Lucey et al.~\cite{lucey2013representing} used the same setup to highlight that a role based assignment of players can eliminate the need of actual player identities in several applications. In basketball, a fixed set of six small cameras are now used for player tracking as a standard in all NBA matches, and the data has been used for extensive analytics~\cite{franks2015counterpoints}. Football certainly has gained the most attention~\cite{gudmundsson2016spatio} and the commercially available data has been utilized for variety of applications from estimating the likelihood of a shot to be a goal~\cite{lucey2014quality} or to learn a team's defensive weaknesses and strengths~\cite{bojinovpressing}.  

The idea of obtaining top view data from broadcast videos has also been explored in previous works, Okuma et al. ~\cite{okuma2004automatic} used KLT~\cite{shi1994good} tracks on manually annotated interest points (with known correspondences) and used them in RANSAC~\cite{fischler1981random} based approach to obtain the homographies in presence of camera pan/tilt/zoom in NHL hockey games. Gupta et al.~\cite{gupta2011using} showed improvement over this work by using SIFT features~\cite{lowe2004distinctive} augmented with line and ellipse information. Similar idea of manually annotating initial frame and then propagating the matches has also been explored in~\cite{lu2013learning}. Li and Chellapa~\cite{li2010group} projected player tracking data from small broadcast clips of American football in top view form to segment group motion patterns. The homographies in their work were also obtained using manually annotated landmarks. 

Hess and Fern~\cite{hess2007improved} build upon~\cite{okuma2004automatic} to eliminate the need of manual initialization of correspondences and proposed an automated method based on SIFT correspondences. Although their approach proposes an improved matching procedure, it may not apply in case of normal football games due to lack of ground visual features. Due to this reason, instead of relying on interest point matches, we move to a more robust edge based approach. Moreover, we use stroke width transforms(SWT)~\cite{epshtein2010detecting} instead of usual edge detectors for filtering out the desired edges. Another drawback of the work in~\cite{hess2007improved} is that the static reference image in their case is manually created, and the process needs to be repeated for each match again. On the other hand, our method is applicable in more generic scenario and we have tested it on data from 16 different matches. The work by Agarwal et al.~\cite{agrawal2015learning} posed the camera transformation prediction between pair of images as a classification problem by binning possible camera movements, assuming that there is a reasonable overlap between the two input images. However, such an approach is not feasible for predicting exact projective transforms. More recently, Homayounfar et. al~\cite{homayounfar2016soccer} presented an algorithm for soccer registration from a single image as a MRF minimization. Their approach relies on vanishing point estimation, which is highly unreliable (in difficult viewpoints, sparse edge detections and shadows). Hence, they have limited their experiments to a small dataset of 105 images to allow manual filtering of vanishing point estimation failures. On the other hand, we experiment on a much thorough dataset (including video sequences).

Our work is also related to camera stabilization method of Grundmann et al.~\cite{grundmann2011auto} which demonstrates that the stabilized camera motion can be represented as combination of distinct constant, linear and parabolic segments. We extend their idea for smoothing the computed homographies over a video. We also benefit from the work of Muja and Lowe~\cite{muja2009fast} for computationally efficient nearest neighbour search.  
%
%
%
%
\section{Method}
The aim of our method is to register a video sequence with a predefined top view static model. The overall framework of our approach is illustrated in Figure~\ref{fig:Overview}. The input image is first pre-processed to remove undesired areas such as crowd and extract visible field lines and obtain a binary edge map. The computed features over this edge map are then used for k-NN search in pre-built dictionary of images with synthetic edge maps and corresponding homographies. Two different stages of smoothing are then performed to improve the video results. We now describe, each of these steps with detail:
%
%
\subsection{Semi supervised dictionary generation}
\label{sec:dictionary_generation}
\begin{figure*}[t]
\centering
\includegraphics[width=0.99\linewidth]{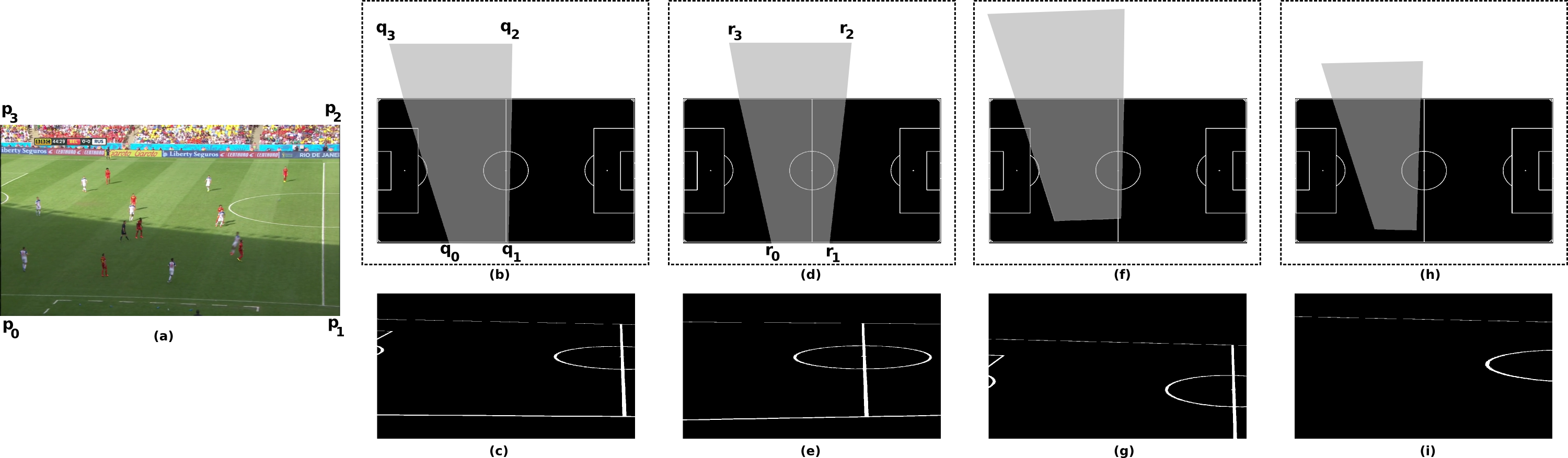}
\caption{Illustration of synthetic dictionary generation. First column shows the input image and second column shows the corresponding registration obtained using manual annotations of point correspondences. The pan, tilt and zoom simulation process is illustrated in third, fourth and fifth column respectively.}
\label{fig:synthetic_dataset_generation}
\end{figure*}
Two images of the same planar surface in space are related by a homography $({\bf H})$. In our case, this relates a given arbitrary image from the football broadcast to the static model of the playing surface. Given a point $x=(u,v,1)$ in one image and the corresponding point $x' = (u',v',1)$, the homography is a $3 \times 3$ matrix, which relates these pixel coordinates $x' = {\bf H}x$. The homography matrix has eight degrees of freedom and can ideally be estimated using 4 pairs of perfect correspondences (giving eight equations). In practice, it is estimated using a RANSAC based approach on a large number of partially noisy point correspondences.  

However, finding a sufficient set of suitable non-collinear candidate point correspondences is difficult in the case of football fields. And manual labelling each frame is not just tedious, it is also challenging task in several images. Due to these reasons, we take an alternate approach: we first hand label the four correspondences in small set of images (where it can be done accurately) and then use them to simulate a large dictionary of `field line images (synthetic edge maps) and related homography pairs'. An example of the process is illustrated in Figure~\ref{fig:synthetic_dataset_generation}. Given a training image (Figure~\ref{fig:synthetic_dataset_generation}(a)), we manually label four points to compute homography $({\bf H_1})$ and register it with the static top view of the ground (Figure~\ref{fig:synthetic_dataset_generation}(b)). We can observe that after applying homography to entire image and warping, the boundary coordinates $(p_0, p_1, p_2, p_3)$ gets projected to points $(q_0,q_1,q_2, q_3)$ respectively. We can now use this to obtain the simulated field edge map (Figure~\ref{fig:synthetic_dataset_generation}(c)) by applying $({\bf H_1^{-1}})$ on the static model (top view). This simulated edge map paired with $ {\bf H_1}$ forms an entry in the dictionary. 

We simulate pan by rotating the quadrilateral $(q_0,q_1,q_2, q_3)$ around the point of convergence of lines $q_0q_3$ and $q_1q_2$ to obtain the modified quadrilateral $(r_0, r_1, r_2, r_3)$, as illustrated in Figure~\ref{fig:synthetic_dataset_generation}(d). Using $(r_0, r_1, r_2, r_3)$ and  $(p_0, p_1, p_2, p_3)$ as respective point correspondences, we can compute the inverse transform $({\bf H_2^{-1}})$ to obtain Figure~\ref{fig:synthetic_dataset_generation}(e). This simulated image along with ${\bf H_2}$ forms another entry in the dictionary. Similarly, we simulate tilt by moving the points $q_0q_3$ and $q_1q_2$ along their respective directions and we simulate zoom by expanding (zoom out) or shinking (zoom-in) the quadrilateral about its center. Now, by using different permutations of pan, tilt and zoom over a set of 75 manually annotated images, we learn a large dictionary $D = \{I_j, H_j\}$ where $I_j$ is the simulated edge map, $H_j$ is corresponding homography and $j \in [0: N-1]$ (we use $N \approx 100K$). We select these 75 images from a larger set of manually annotated images, using a weighted sampling from a hierarchical cluster (using the $H$ matrix as feature for clustering). The permutations of pan, tilt, zoom were chosen carefully to comprehensively cover the different field of views. We can observe that the proposed algorithm is able to generate viewpoint homography pair like Figure~\ref{fig:synthetic_dataset_generation}(i)), which may be infeasible to get using manual annotation (due to lack of distinctive points).  
\subsection{Nearest neighbour search algorithms}
\label{sec:NN_algorithms}
We pose the homography estimation problem as the nearest neighbour search over the synthetic edge map dictionary. Given a preprocessed input image and its edge map $x$, we find the best matching edge map $I_j$ (or $k$ best matching edge maps) from the dictionary and output the corresponding homography $H_j$ (or set of $k$ homographies). In this section, we present three different approaches we anlayzed for computing the nearest neighbours. We specifically choose an image gradient based approach (HOG), a direct contour matching approach (chamfer matching) and an approach learning abstract mid level features (CNN's). 
\subsubsection{Chamfer matching based approach}
\begin{figure}[t]
\centering
\includegraphics[width=0.99\linewidth]{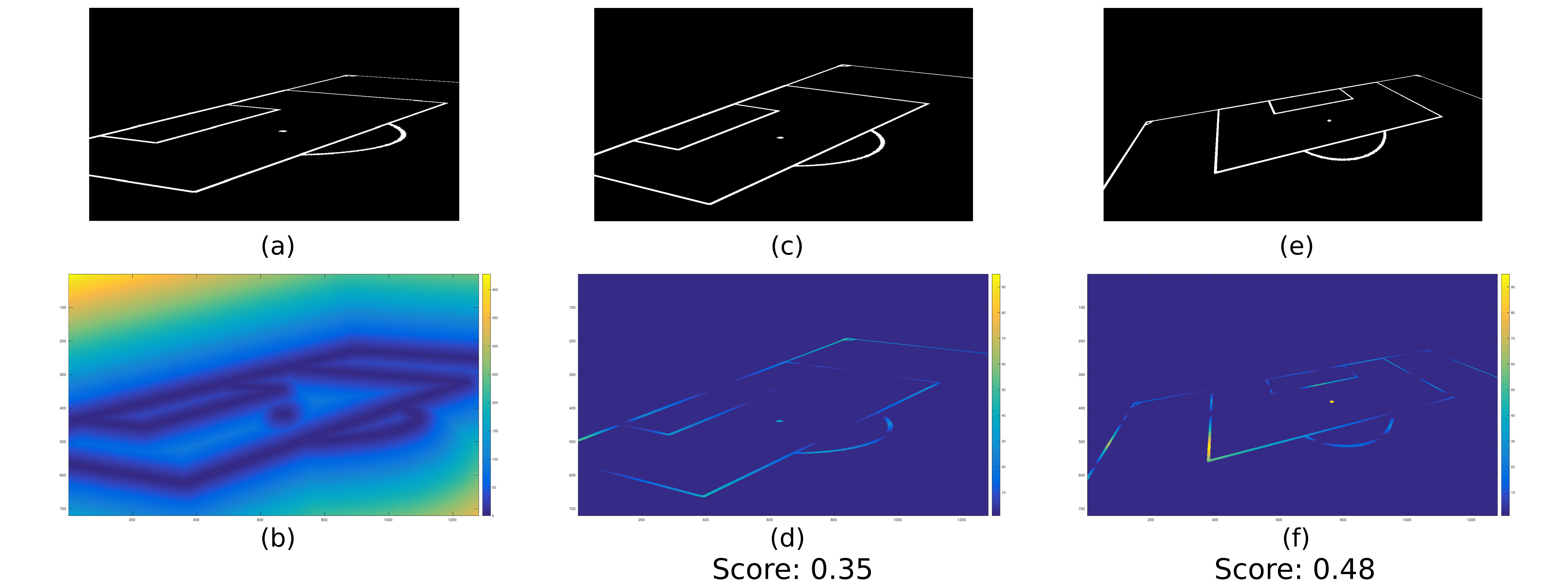}
\caption{Illustration of chamfer matching. The first column shows the input image $x$ and its distance transform $T(x)$. The second  and third column show two different edge maps and their multiplication with $T(x)$. We can observe that image (c) is a closer match and gives a lower chamfer distance. }
\label{fig:chamfer}
\end{figure}
The first method we propose is based on chamfer matching~\cite{barrow1977parametric}, which is a popular technique to find the best alignment between two edge maps. Although proposed decades ago, it remains a preferred method for several reasons like speed and accuracy, as discussed in~\cite{thayananthan2003shape}. Given two edge maps $x$ and $I_j$, the chamfer distance quantifies the matching between them. The chamfer distance is the mean of the distances between each edge pixel in $x$ and its closest edge pixel in $I_j$. It can be efficiently computed using the distance transform function $T(.)$, which takes a binary edge image as input and assigns to each pixel in the image the distance to its nearest edge pixel. The chamfer matching then reduces to a simple multiplication of the distance transform on one image with the other binary edge image. The process is illustrated in Figure~\ref{fig:chamfer}.  
We use the chamfer distance for the nearest neighbour search. Given an input image $x$ and its distance transform $T(x)$ we search for index $j^*$ in the dictionary, such that
\begin{equation}
\small
j^{*} = \underset{j}{\mathrm{argmin}} \frac{T(x) . I_j}{\| I_j \|_{1}} \mbox{ ,}
\end{equation}
where $\| \|_{1}$ is the $\ell_1$ norm and the index $j^*$ gives the index of the true nearest neighbour. Given an epsilon $\epsilon>0$, the approximate nearest neighbours are given by list of indices $j$, such that $\frac{T(x) . I_{j}}{\| I_{j} \|_{1}} \le (1+\epsilon)\frac{T(x) . I_{j^*}}{\| I_{j^*} \|_{1}}$.

\subsubsection{HOG based approach}
The second method is based on HOG features~\cite{dalal2005histograms}, where the nearest neighbour search is performed using the euclidean distance on the HOG features computed over both the dictionary edge maps and the input edge map. So, given the input edge map $x$ and its corresponding HOG features $\phi_h(x)$ we search for $j^*$ in the dictionary, such that
\begin{equation}
\small
j^{*} = \underset{j}{\mathrm{argmin}} \| \phi_h(x) - \phi_h(I_j) \|_2 \mbox{ ,}
\end{equation}
where $\| \|_{2}$ is the $\ell_2$ norm. 
\subsubsection{CNN based approach}
%
It has been shown that CNN features learnt for one task like object classification, can be efficiently used for other tasks like object localization ~\cite{oquab2014learning}. On the similar lines, we use the mid level features learnt using the network architecture of Qian et al.~\cite{yu2015sketch} and Krizhevsky et al~\cite{krizhevsky2012imagenet}. The architecture in~\cite{yu2015sketch} is trained for sketch classification and AlexNet~\cite{krizhevsky2012imagenet} has been trained for ImageNet~\cite{russakovsky2015imagenet}. We remove the last fully connected layer in both cases and use it as the feature vector for the nearest neighbour search. 

Given the input edge map $x$ and its output at last fully connected layer $\phi_c(x)$ we search for $j^*$ in the dictionary, such that
\begin{equation}
\small
j^{*} = \underset{j}{\mathrm{argmin}} \| \phi_c(x) - \phi_c(I_j) \|_2 \mbox{ ,}
\end{equation}
where $\| \|_{2}$ is the $\ell_2$ norm.        
\subsection{Smoothing and Stabilization}
\label{sec:smoothing_and_stabilization}
For a given input video sequence, we compute $k$ homography candidates independently for each frame using the nearest neighbour search algorithms described above. Just taking the true nearest neighbour for each frame independently may not always give the best results due to noise in the pre-processing stage or the absence of a close match in the simulated dictionary. To remove outliers and to obtain a jerk free and stabilized camera projections, we use two different optimization stages. The first stage uses a markov random field (MRF) based optimization, which selects one of the $k$ predicted homographies for each frame to remove the outliers and discontinuities. The second stage further optimizes these discrete choices, to obtain a more smooth and stabilized camera motion. 
\subsubsection{MRF optimization}   
The algorithm takes as input the $k$ predicted homographies for each frame with their corresponding nearest neighbour distances and outputs a sequence of $\xi = {\{s_{t}\}}$ states $s_t \in [1: k] $, for all frames $t = [1:N]$. It minimizes the following global cost function: 
\begin{equation}
\small
E(\xi) = \sum_{t=1}^{N} E_{d}(s_{t}) + \sum_{t=2}^{N} E_{s}(s_{t-1},s_{t}).
\label{eq:dynamic_programming}
\end{equation}
The cost function consists of a data term $E_{d}$ that measures the evidence of the object state using the nearest neighbour distances and a smoothness term $E_{s}$ which penalizes sudden changes. The data term and the smoothness term are defined as follows:
\begin{equation}
\small
 E_{d}(s_{t}) =  \log (P(s_{t},t))\mbox{ .}
\end{equation}
Here, $P(s_{t},t)$ is the nearest neighbour distance for state $s_{t}$ at frame $t$. And 

\begin{equation}
\small
 E_{s}(s_{t-1},s_{t}) = \| H_{s_t} - H_{s_{t-1}} \|_2, 
\end{equation}
is the Euclidean distance between the two ($3 \times 3$) homography matrices, normalized so that each of the eight parameters lie in a similar range. Finally, we use dynamic programming (DP) to solve the optimization problem presented in Equation~\ref{eq:dynamic_programming}.
%
%
\subsubsection{Camera stabilization}
\begin{figure*}[t]
\centering
\begin{tabular}[b]{c c c c c}
\includegraphics[width=0.18\linewidth]{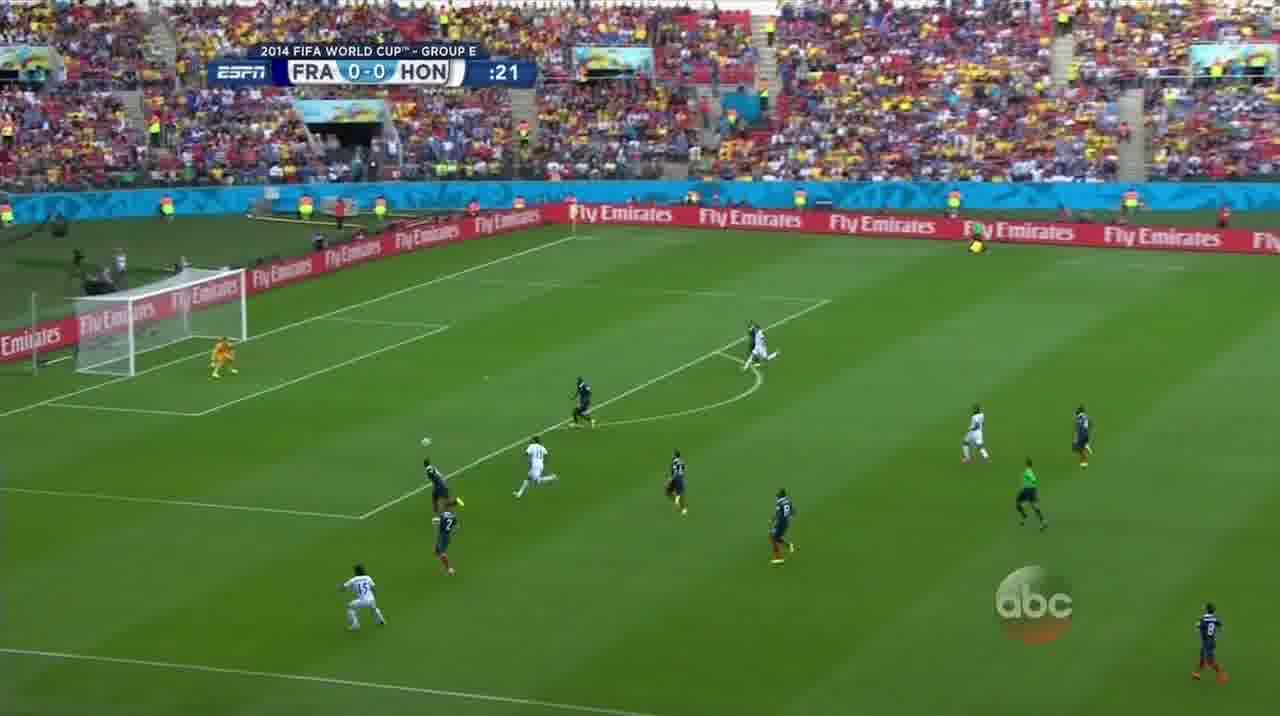} &\hspace{-1em}
\includegraphics[width=0.18\linewidth]{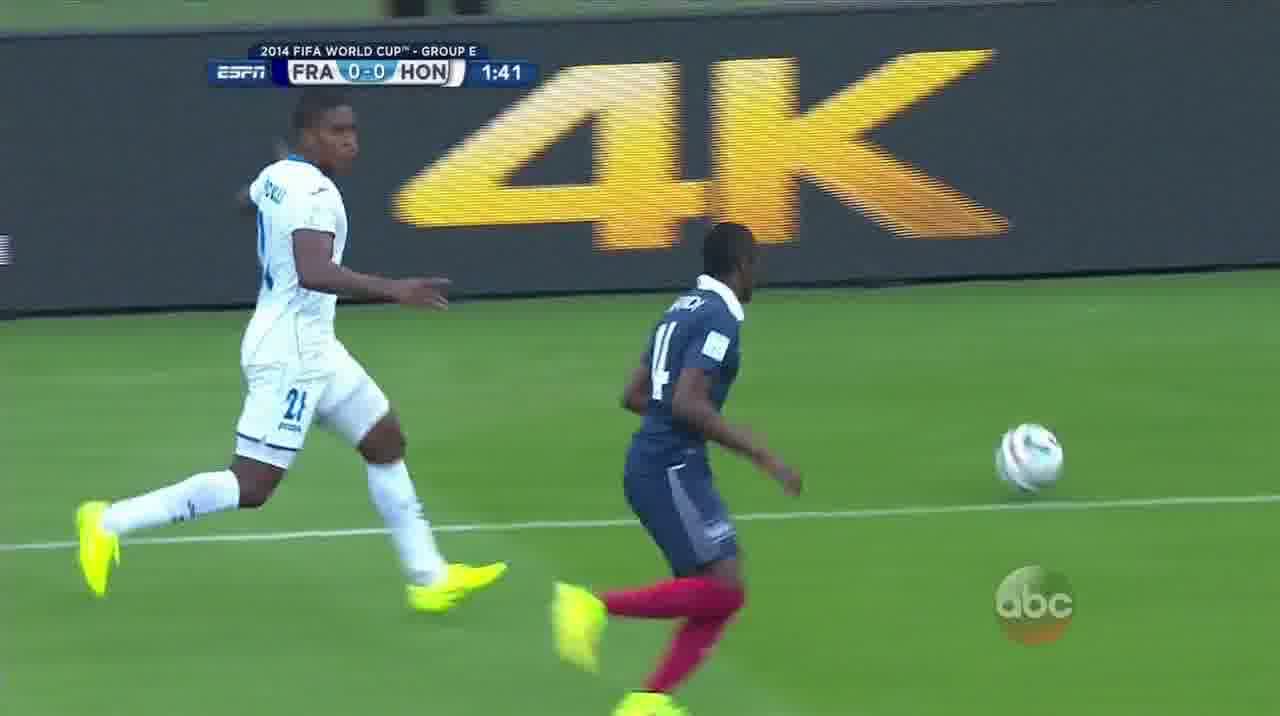} &\hspace{-1em}
\includegraphics[width=0.18\linewidth]{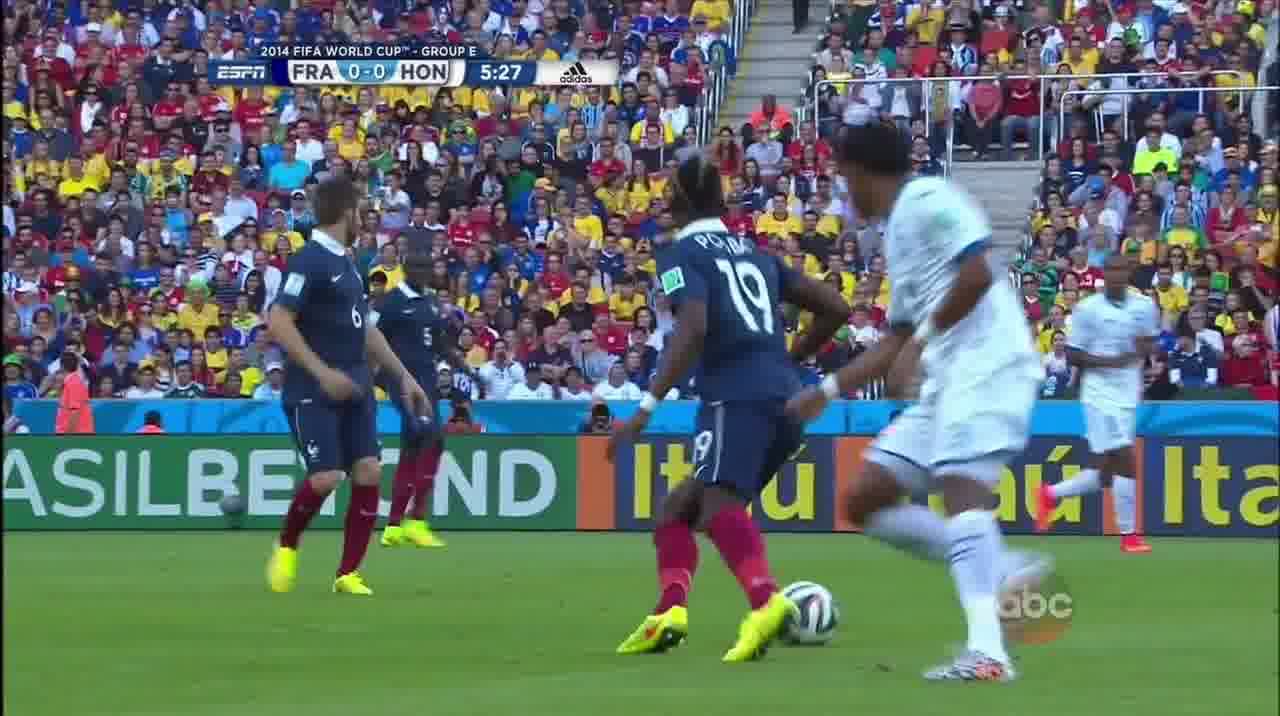} &\hspace{-1em}
\includegraphics[width=0.18\linewidth]{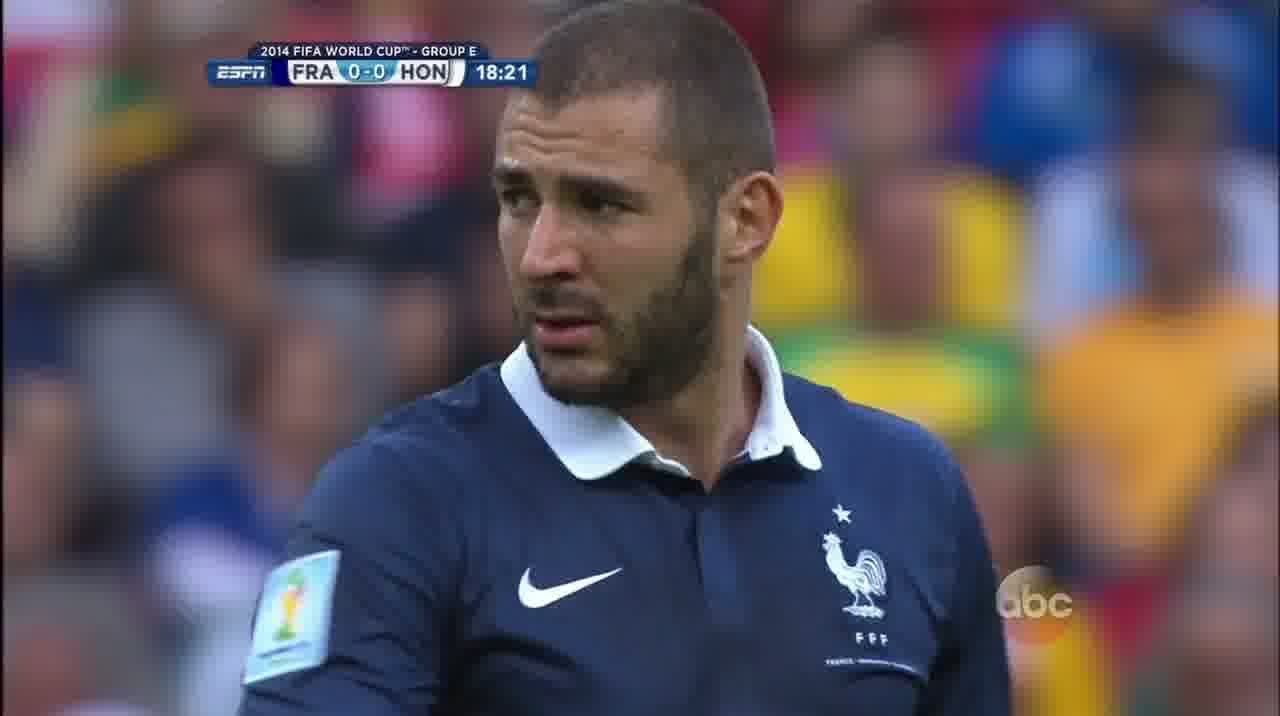}& \hspace{-1em}
\includegraphics[width=0.18\linewidth]{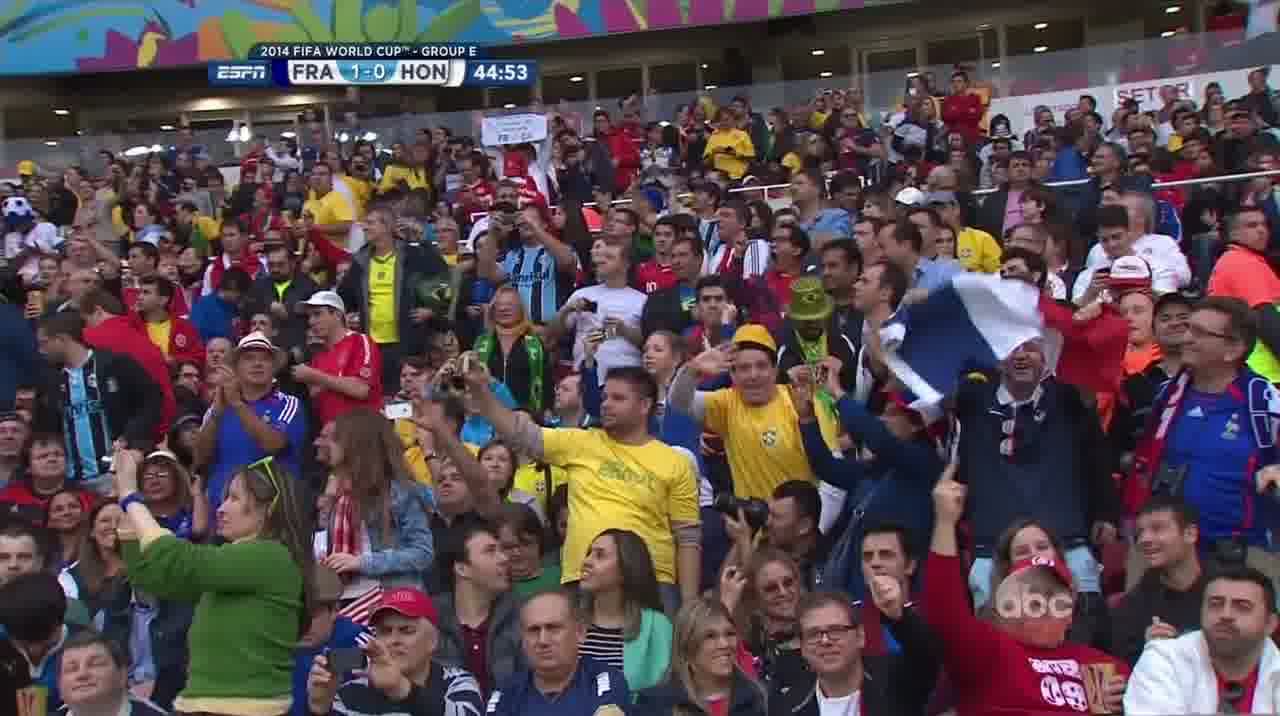} \\
\end{tabular}
\caption{We classify the camera viewpoints from a usual football broadcast into five different categories namely (from left to right) top zoom-out, top zoom-in, ground zoom-out, ground zoom-in and miscellaneous (covering mainly the crowd view).}
\label{fig:camera_labels}
\end{figure*}
The MRF optimization removes the outliers and the large jerks, however a small camera jitter still remains because its output is a discrete selection at each frame.  We solve this problem using a solution inspired by the previous work on camera stabilization~\cite{grundmann2011auto}. The idea is to break the camera trajectory into distinct constant (no camera movement), linear (camera moves with constant velocity) and parabolic (camera moves with constant acceleration or deceleration) segments.  We found that this idea also correlates with the camera work by professional cinematographers, who tend to keep the camera constant as much as possible, and when the movement is motivated they constantly accelerate, follow the subject (constant velocity) and then decelerate to static state~\cite{gramofedit}. The work in ~\cite{grundmann2011auto} shows that this can be formalized as a L1-norm optimization problem. 

However the idea of~\cite{grundmann2011auto} cannot be directly applied in our case, as we can not rely on interest point features for the optimization, because we are already in projected top view space. We parametrize the projected polygon (for example the quadrilateral $q_0q_1q_2q_3$ in Figure~\ref{fig:synthetic_dataset_generation}) using six parameters, the center of the camera $(cx, cy)$, the pan angle $\theta$, the zoom angle $\phi$ and two intercepts $(r1, r2)$ (for near clipping plane and far clipping plane respectively). Given a video of $N$ frames, we formulate the stabilization as convex optimization over the projected plane $P_t = \{ cx_t,cy_t,\theta_t,\phi_t,r1_t,r2_t \}$ at each frame $t \in [0:N-1] $. We solve for $P^*_t$ which minimizes the following energy function:
\begin{equation}
\footnotesize
\begin{split}
E_c =  \sum_{t=1}^{N} (P^*_t - P_t)^2 + \lambda_1\sum_{t=1}^{N-1} \| P^*_{t+1} - P^*_{t} \|_1  \\+  \lambda_2\sum_{t=1}^{N-2} \| P^*_{t+2} - 2P^*_{t+1}  + P^*_{t} \|_1 \\
+ \lambda_3\sum_{t=1}^{N-3} \| P^*_{t+3} - 3P^*_{t+2} + 3P^*_{t+1} - P^*_{t} \|_1. 
\end{split}
\end{equation}
The energy function $E_c$ comprises of a data term and three L1-norm terms over the first order, second order and the third order derivatives and $\lambda_1$, $\lambda_2$ and $\lambda_3$ are parameters. As $E_c$ is convex, it can be efficiently solved using any off the shelf solver, we use cvx~\cite{cvx}.   
\section{Experimental Results}
\begin{figure*}[t]
\centering
\includegraphics[width=0.9\linewidth]{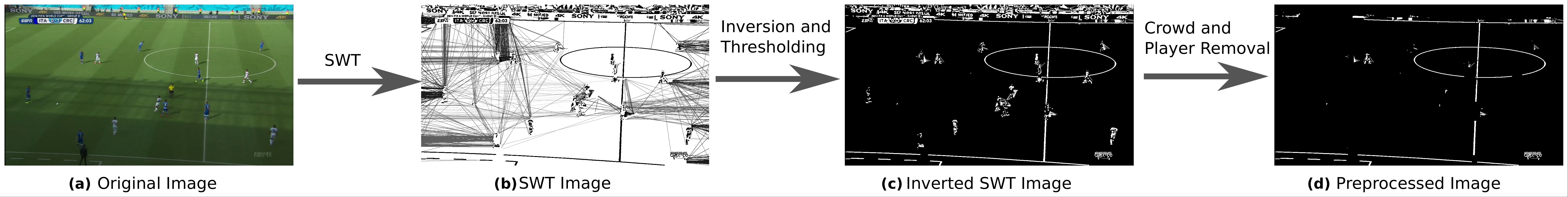}
\caption{Illustration of the pre-processing pipeline. Observe that how SWT is able to filter out the field lines in presence of complex shadows (usual edge detectors will fail in such scenarios).}
\label{fig:preprocessing}
\end{figure*}
We perform experiments on broadcast images and video sequences selected from 16 matches of football world cup 2014. We evaluate our work using three different experiments. The first experiment compares the three matching approaches (chamfer, HOG and CNN based) over a large simulated test dataset. The second experiment draws similar comparison over actual broadcast images from different matches with different teams in varying conditions. The third experiment showcases the results over broadcast video sequences, comparing with previous methods~\cite{okuma2004automatic,lu2013learning} and highlighting the benefits of the camera smoothing and stabilization. 
%
\subsection{Results over simulated edge maps}
Similar to the procedure explained in section~\ref{sec:dictionary_generation}, we generate a set of 10000 edge map and homography pairs and use it as a test dataset. We annotated a different set of images for generating this test dataset to keep it distinct with the training set (used for learning the dictionary). Then, we compute the nearest neighbour using the three approaches explained in section~\ref{sec:NN_algorithms} on each of the test image (edge map) independently. We use the computed homographies to project the given test image over the static model and obtain a polygon $P_e$.  Since, the simulated dataset also contains the corresponding ground truth homography matrix, we then use it to obtain actual ground truth top view estimation, which gives another polygon $P_g$. To evaluate, we use the intersection-over-union (IOU) measure over the ground truth and the estimated polygons i.e. $\frac{P_e \cap P_g}{P_e \cup P_g}$ (also known as Jaccard index). 

The results are illustrated in Table~\ref{table:simulated_results}. Interestingly, all methods give a mean IOU measure above 80\%, with HOG and SketchNet based features crossing 90\%. Since, the intersection-over-union measure decreases quite rapidly, a 90\% accuracy shows that the idea works nearly perfect in absence of noise with these features. Moreover, the high median IOU measure suggests that most images are accurately registered. 
%
%
\setlength{\tabcolsep}{5pt}
\setlength{\extrarowheight}{1pt}
\begin{table}[t!]
\scriptsize
\begin{center}
\begin{tabular}{lll}
\hline 
\multicolumn{3}{c}{{\bf Synthetic Dataset}} \\ \hline \setlength{\extrarowheight}{1.5pt}
 & Mean & Median \\
NN-Chamfer & 83.2 & 89.2\\
NN-HOG & 90.9 & 92.4 \\
NN-AlexNet & 88.4 & 90.7\\ 
NN-SketchNet & 93.1 & 94.4\\ 
\hline \\
\end{tabular} \hspace{1mm} \setlength{\extrarowheight}{1pt}
\begin{tabular}{lll}
\hline 
\multicolumn{3}{c}{{\bf Broadcast image dataset}} \\ \hline \setlength{\extrarowheight}{1.5pt}
 & Mean & Median \\
NN-Chamfer & 80.5 & 83.2\\
NN-HOG & 85.8 & 88.9\\
NN-AlexNet & 66.1 & 69.3\\ 
NN-SketchNet & 14.1 & 0.0 \\
\hline \\
\end{tabular}
\caption{Results over the synthetically generated test dataset (left) and results over the real broadcast image dataset (right).}
\label{table:simulated_results}
\end{center}
\end{table}
\subsection{Results over broadcast images}
The proposed method can only be practically applicable if it can broadly replicate the accuracy obtained on synthetic dataset over sampled RGB images from broadcast videos. Since, the nearest neighbour search takes as input the features over edge maps, we need to first pre-process the RGB images to obtain the edge maps (only containing the field lines). Moreover, a football broadcast consists of different kind of camera viewpoints (illustrated in Figure~\ref{fig:camera_labels}) and the field lines are only properly visible in the far top zoom-out view (which though covers nearly seventy five percent of the broadcast video frames). Henceforth, we propose a two stage pre-processing algorithm:
\subsubsection{Pre-processing}
The first pre-processing step selects the top zoom-out frames from a given video sequence. We employ the classical Bag of Words (BoW) representation on SIFT features to classify each frame into one of the five classes illustrated in Figure~\ref{fig:camera_labels}. We use a linear SVM to perform per frame classification (taking features from a temporal window of 40 frames centred around it), followed by a temporal smoothing. Even using this simple approach, we achieve an accuracy of 98 percent, for the top zoom-out class label (trained over 45 minutes of video and tested over 45 minutes of video from another match).  

Now, given the top zoom-out images from the video, the second pre-processing step extracts the edge map with field lines. The entire procedure is illustrated in Figure~\ref{fig:preprocessing}. First we compute the stroke width transform (SWT) over the input images and filter out the strokes of size more than 10 pixels (preserving the field lines which comprise of small consistent stroke widths). The benefit of using SWT over usual methods like canny edge detection is that it is more robust to noise like shadows, field stripes (light-dark stripes of green colors) etc. We further remove crowd (using color based segmentation of field) and players (using Faster-RCNN human detector~\cite{renNIPS15fasterrcnn}) to obtain the edge map, primarily containing only the field lines with partial noise(Figure~\ref{fig:preprocessing}(d)). 
\subsubsection{Quantitative evaluation}
\begin{figure}[t]
\centering
\begin{tabular}[b]{c c c c c}
\hspace{-1em} \includegraphics[width=0.24\linewidth]{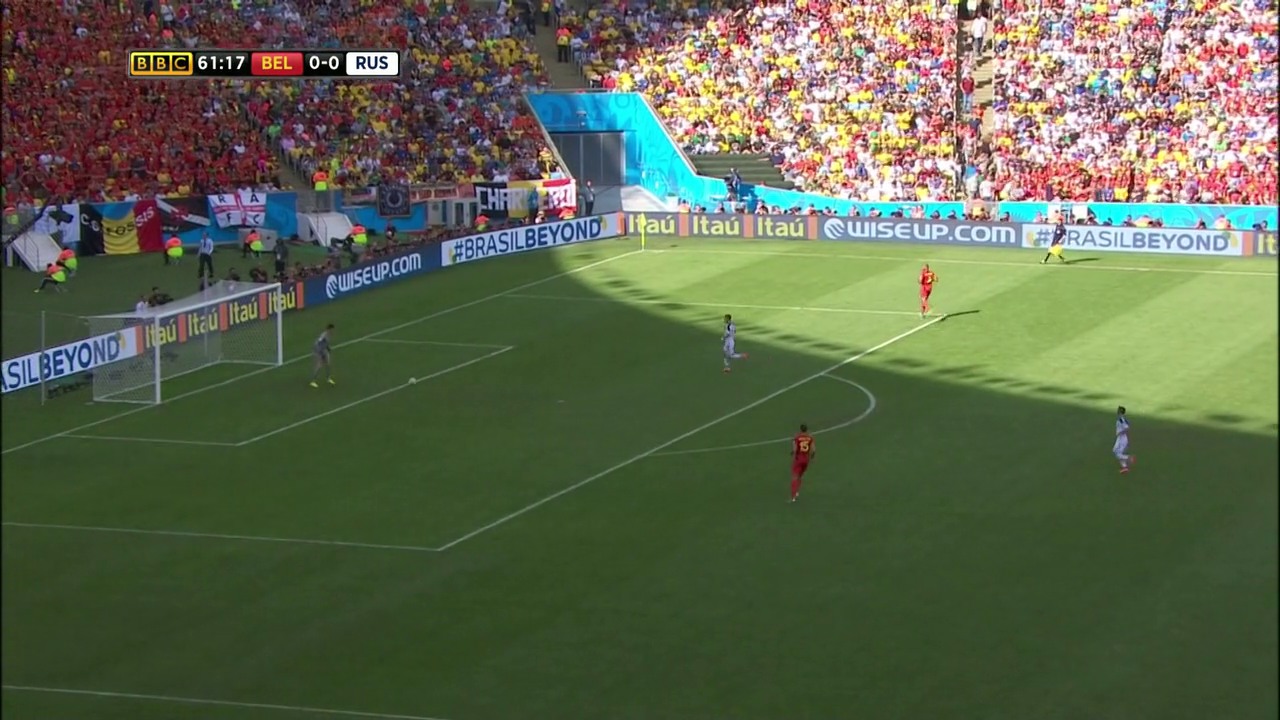} &\hspace{-1.1em}
\includegraphics[width=0.24\linewidth]{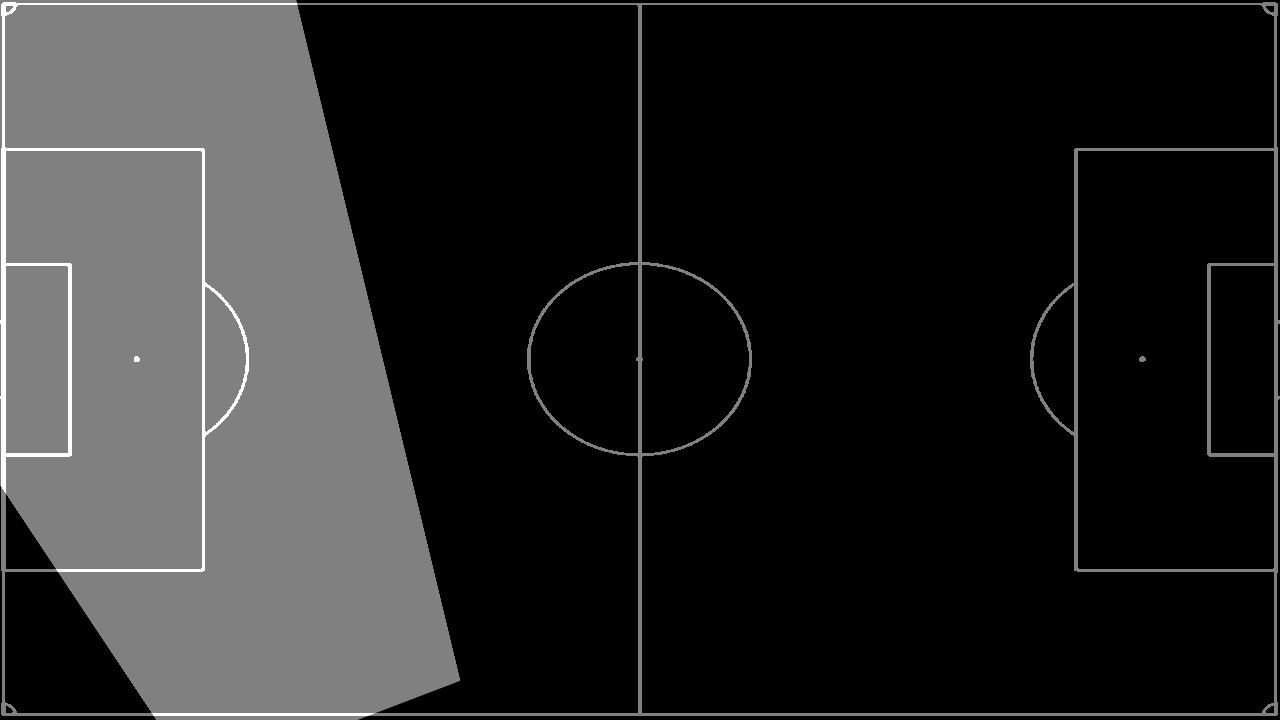} &\hspace{-1em}
\includegraphics[width=0.24\linewidth]{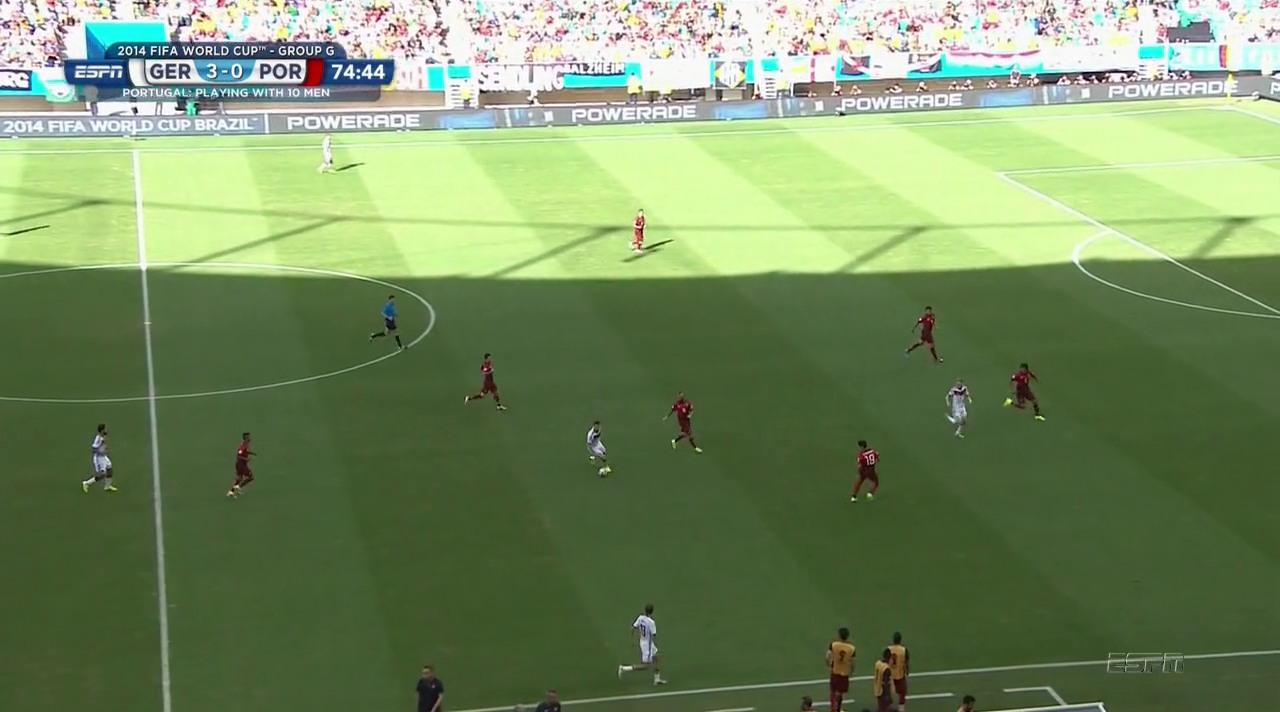} &\hspace{-1.1em}
\includegraphics[width=0.24\linewidth]{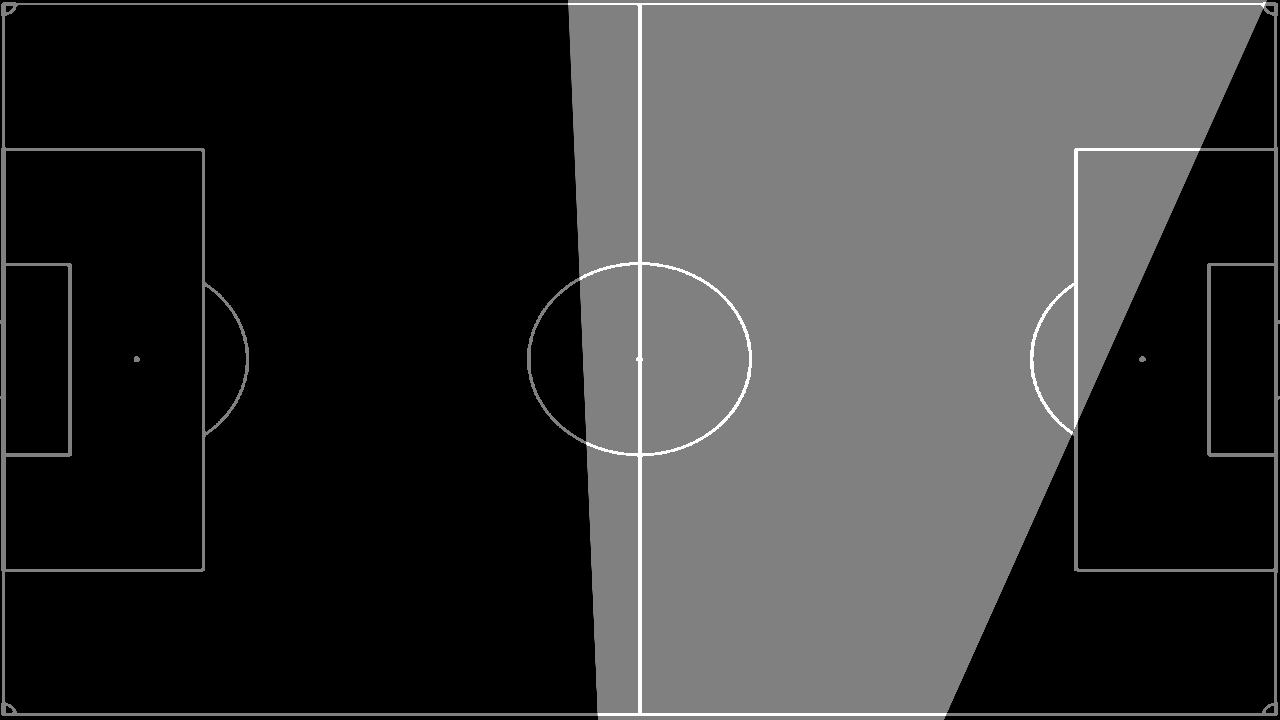} &\hspace{-1em} \vspace{-2.7mm}\\ 
\multicolumn{2}{c}{\tiny (a)} & \multicolumn{2}{c}{\tiny (b)} \vspace{0.5mm}\\
\hspace{-1em}\includegraphics[width=0.24\linewidth]{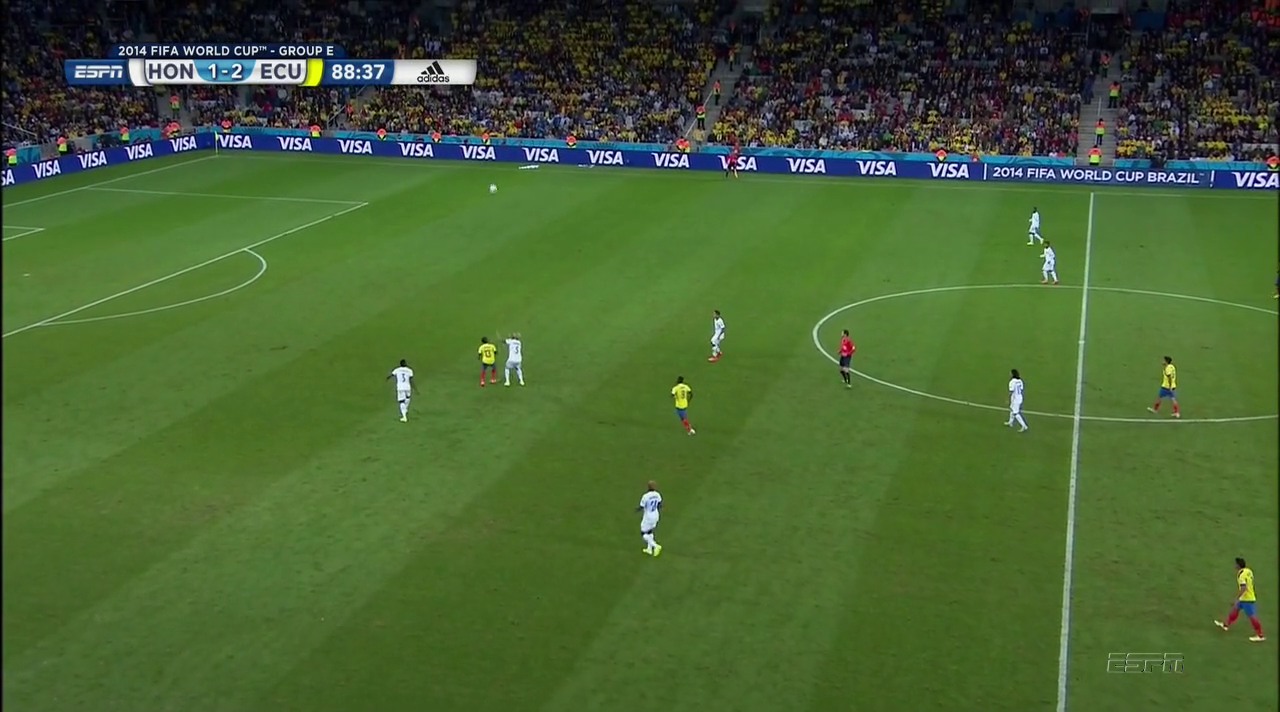} &\hspace{-1.1em}
\includegraphics[width=0.24\linewidth]{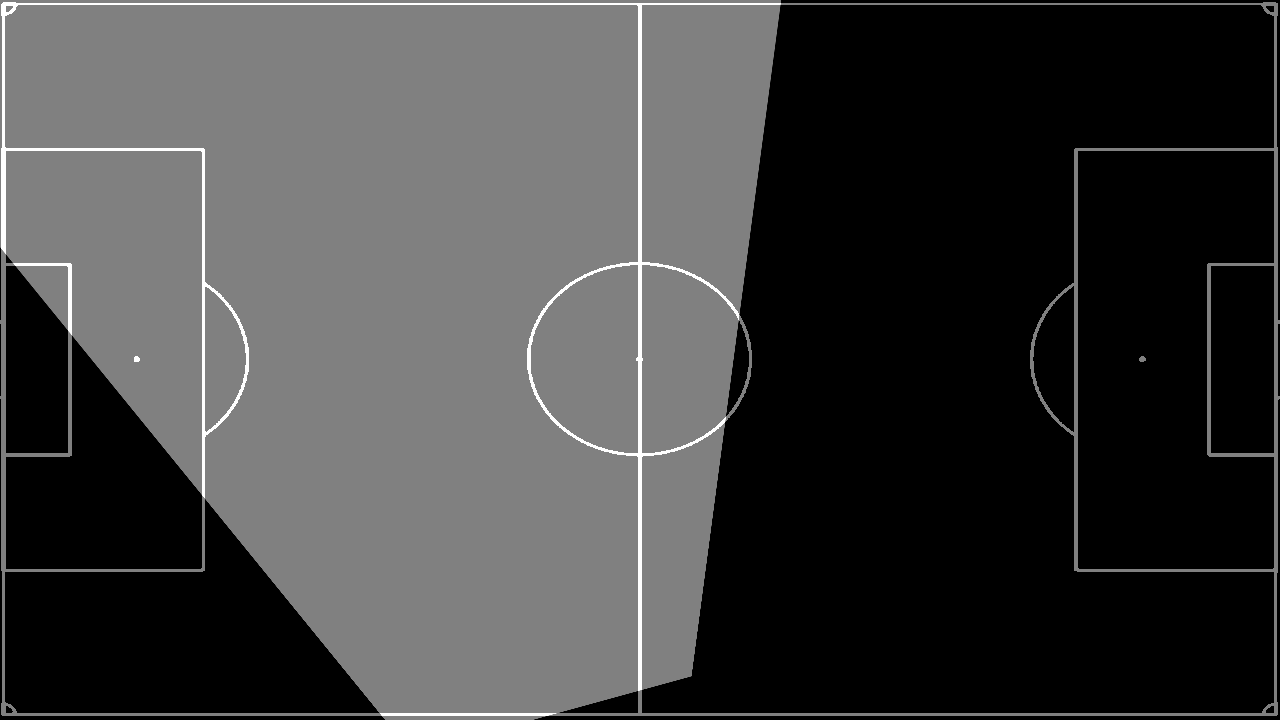} &\hspace{-1em}
\includegraphics[width=0.24\linewidth]{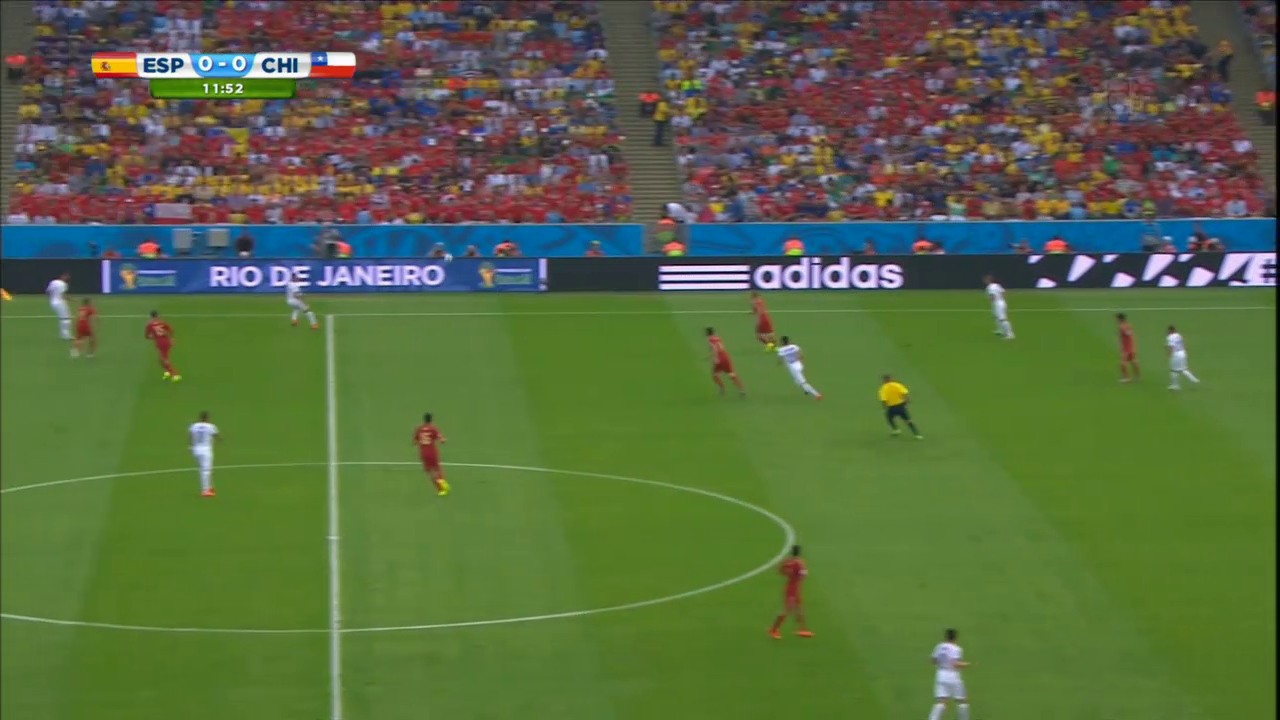} &\hspace{-1.1em}
\includegraphics[width=0.24\linewidth]{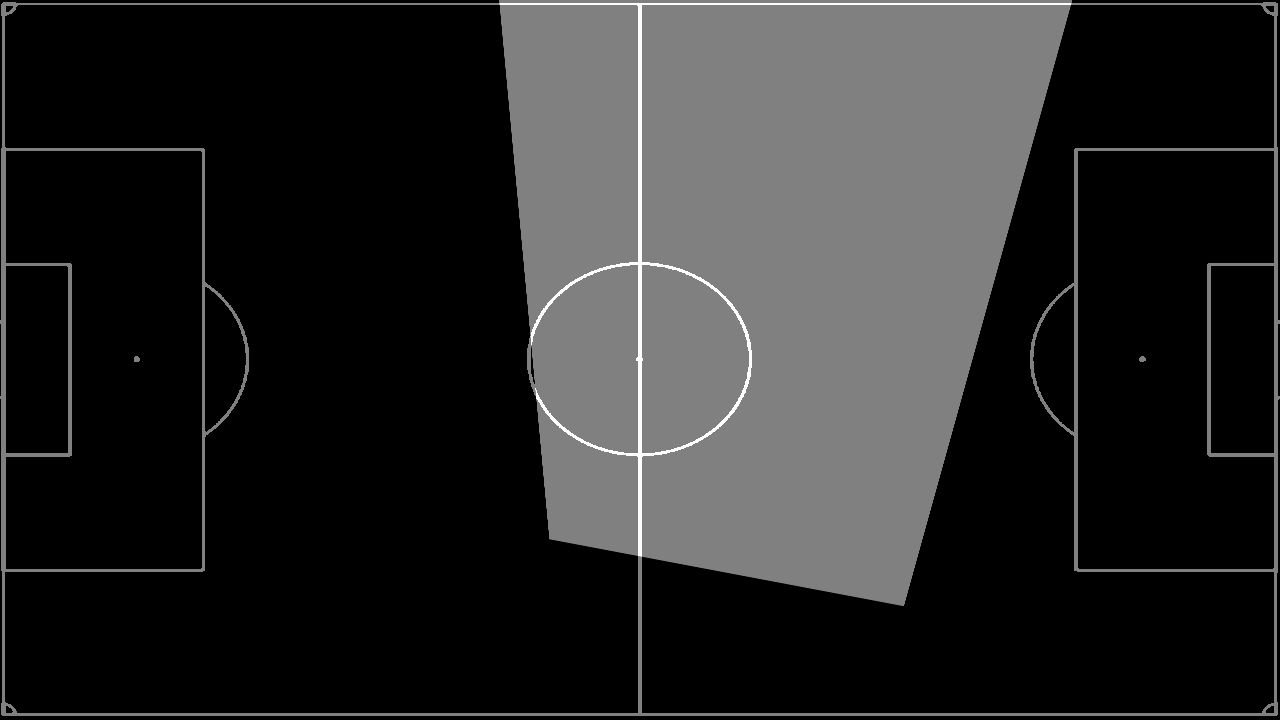} &\hspace{-1em} \vspace{-2.7mm} \\
\multicolumn{2}{c}{\tiny (c)} & \multicolumn{2}{c}{\tiny (d)} \vspace{0.5mm}\\
\hspace{-1em}\includegraphics[width=0.24\linewidth]{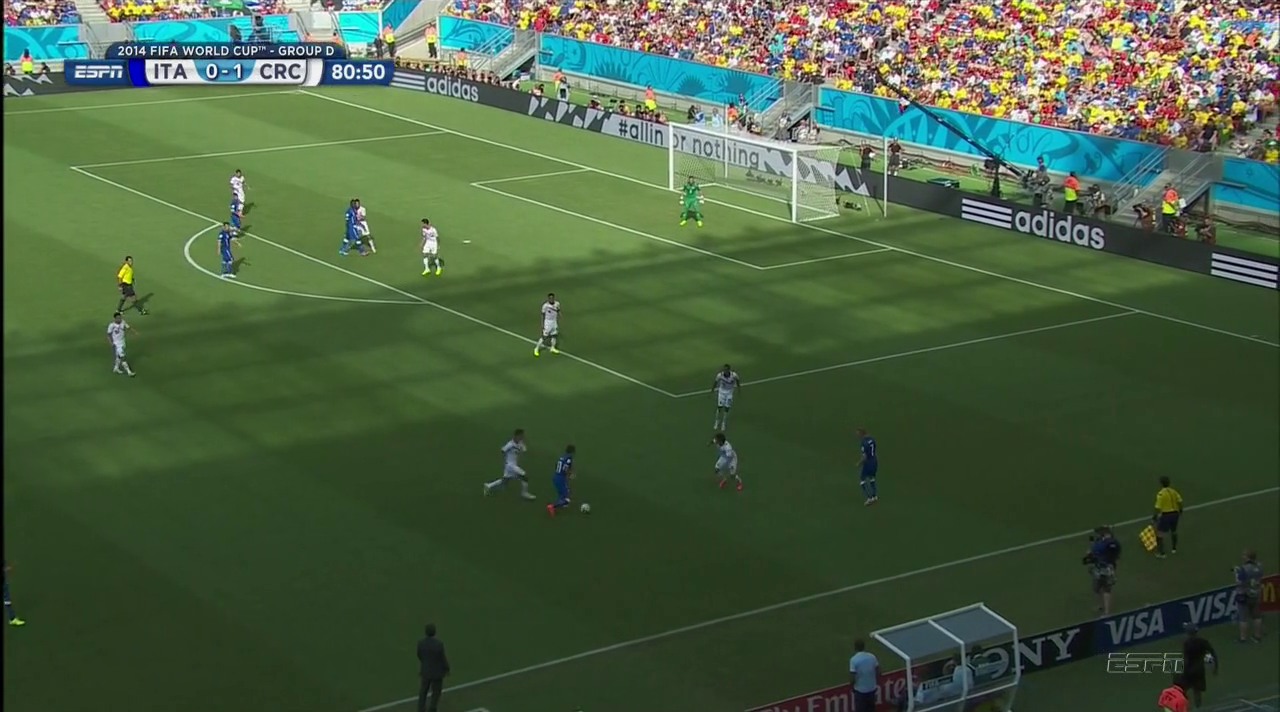} &\hspace{-1.1em}
\includegraphics[width=0.24\linewidth]{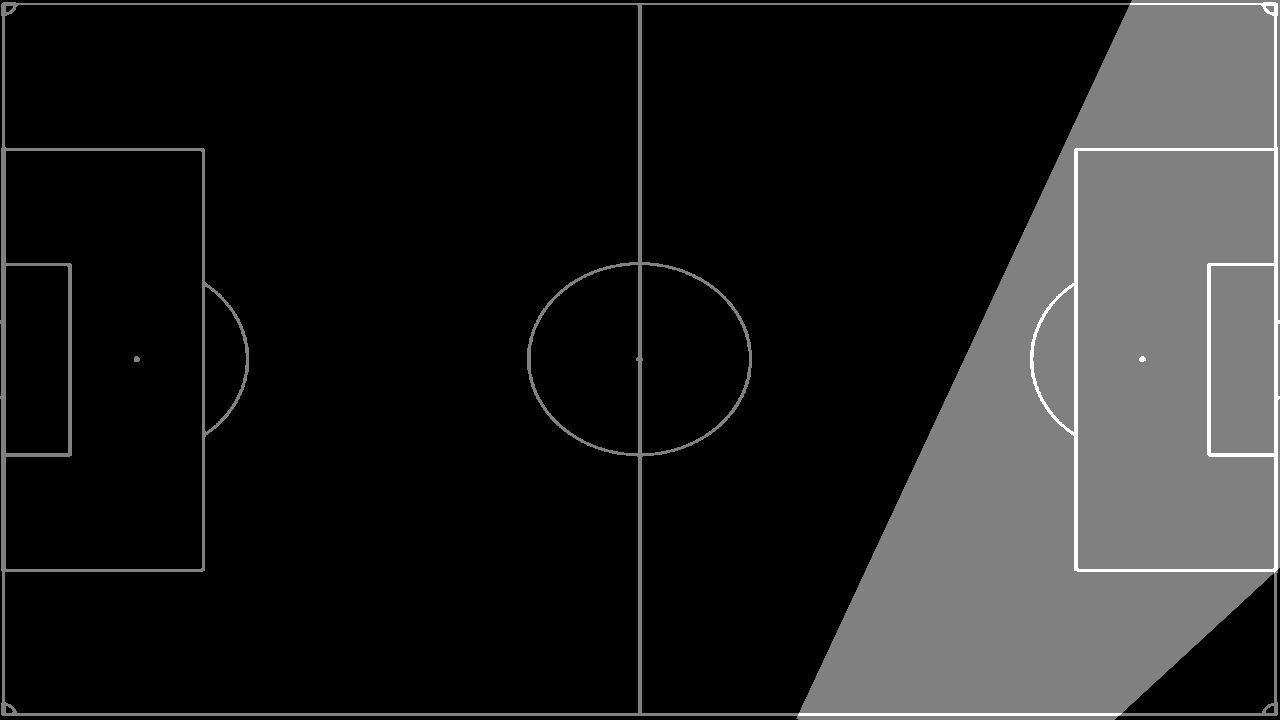} &\hspace{-1em}
\includegraphics[width=0.24\linewidth]{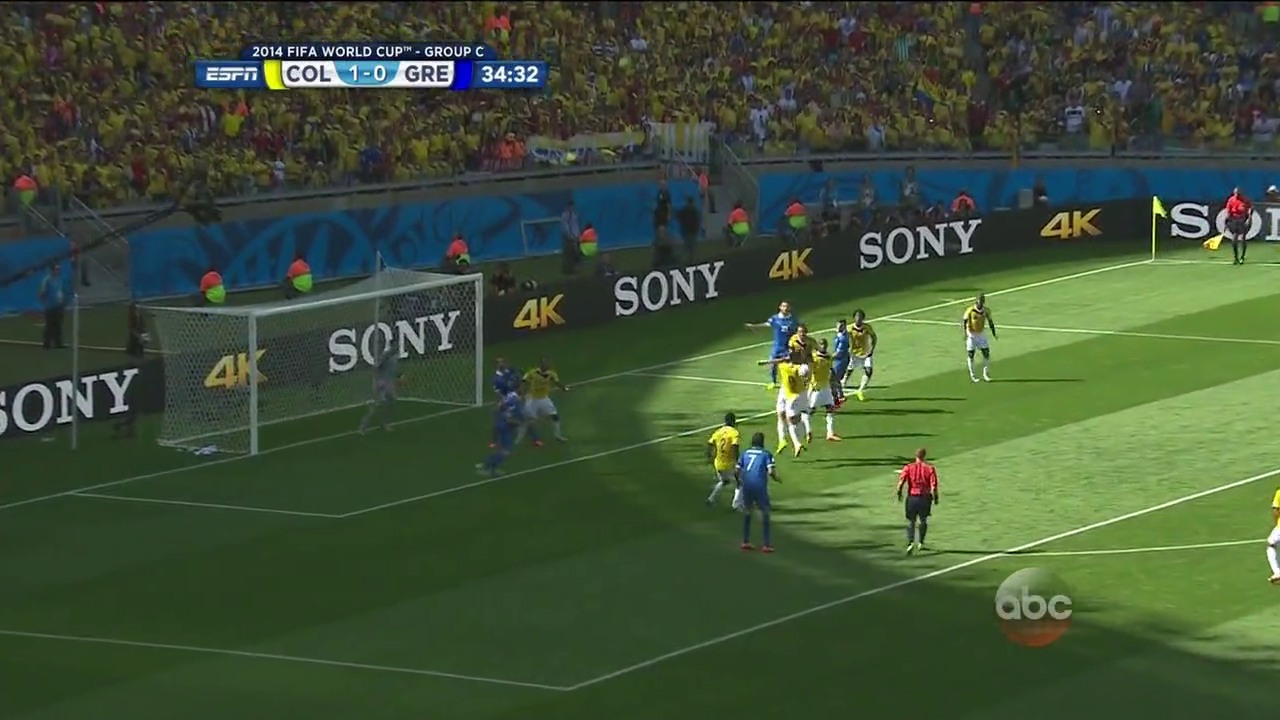} &\hspace{-1.1em}
\includegraphics[width=0.24\linewidth]{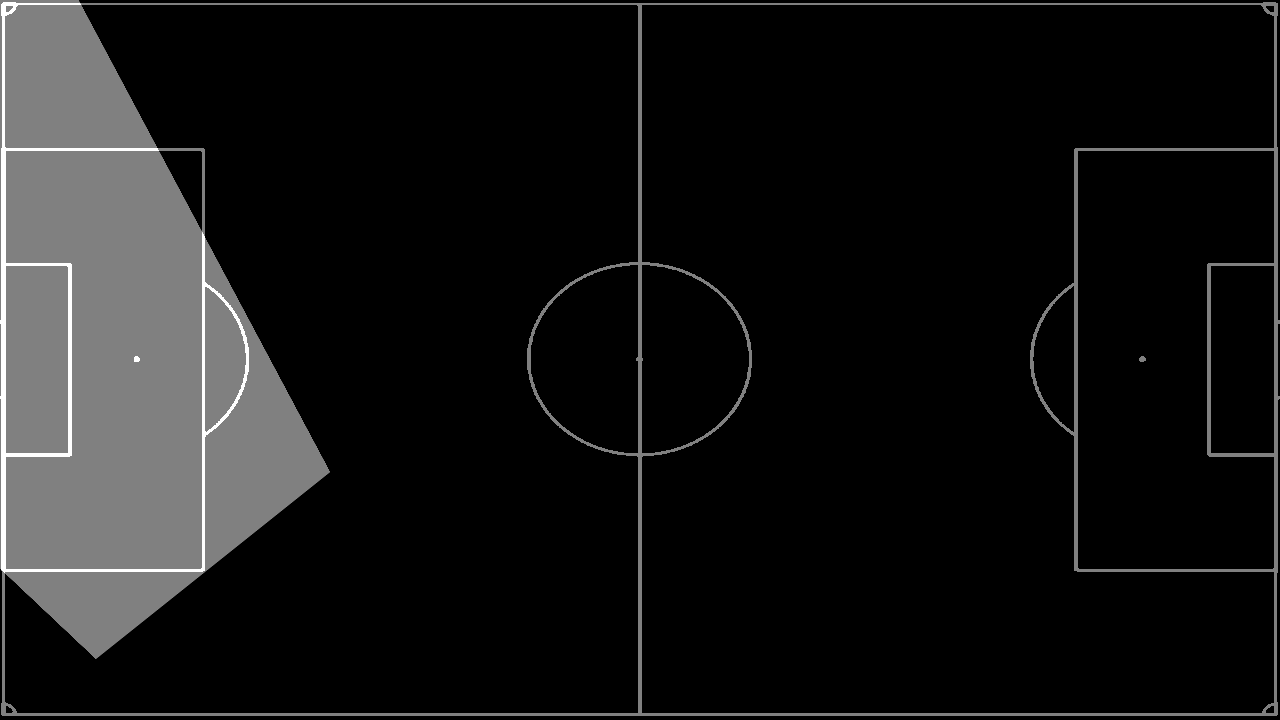} &\hspace{-1em} \vspace{-2.7mm}\\
\multicolumn{2}{c}{\tiny (e)} & \multicolumn{2}{c}{\tiny (f)} \vspace{0.5mm}\\
\hspace{-1em}\includegraphics[width=0.24\linewidth]{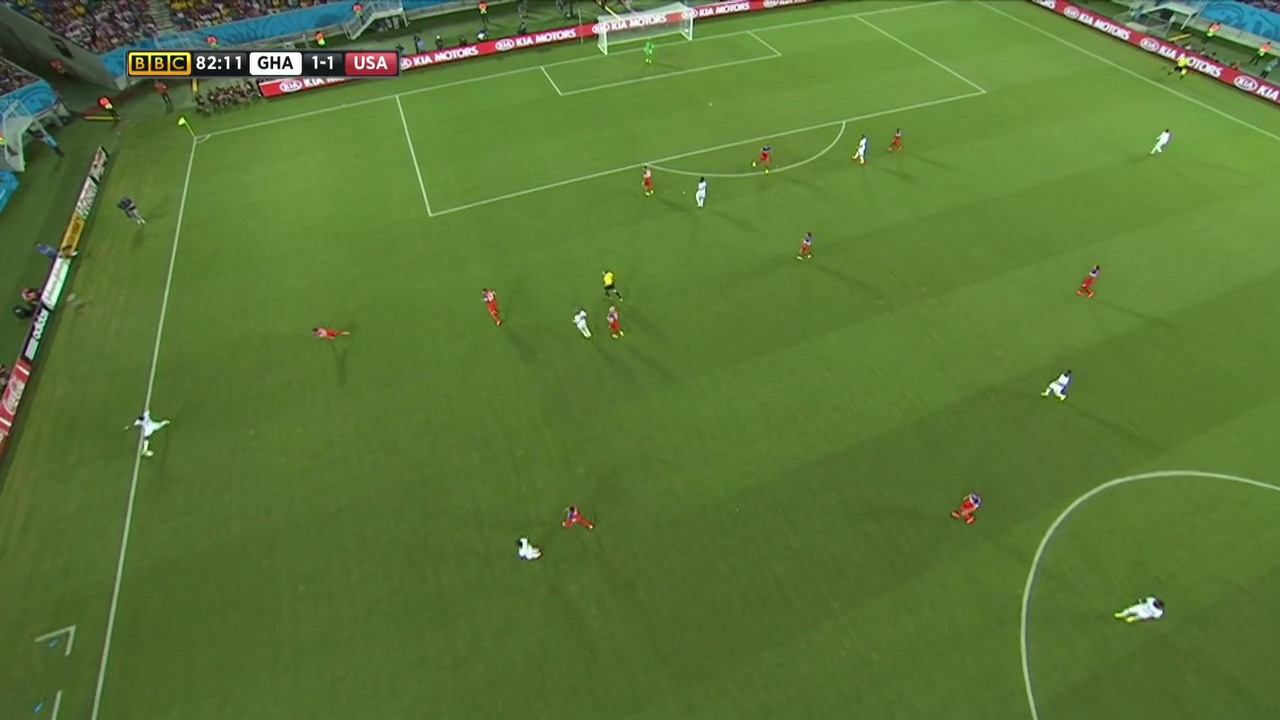} &\hspace{-1.1em}
\includegraphics[width=0.24\linewidth]{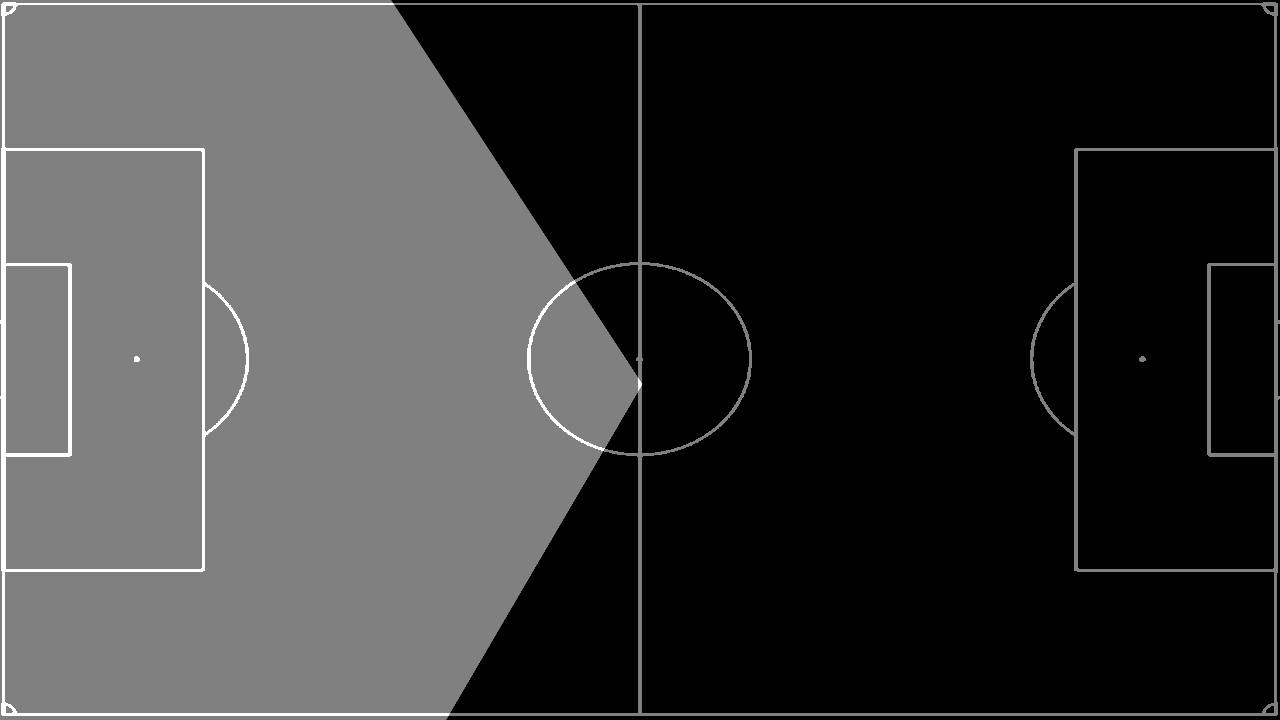} &\hspace{-1em}
\includegraphics[width=0.24\linewidth]{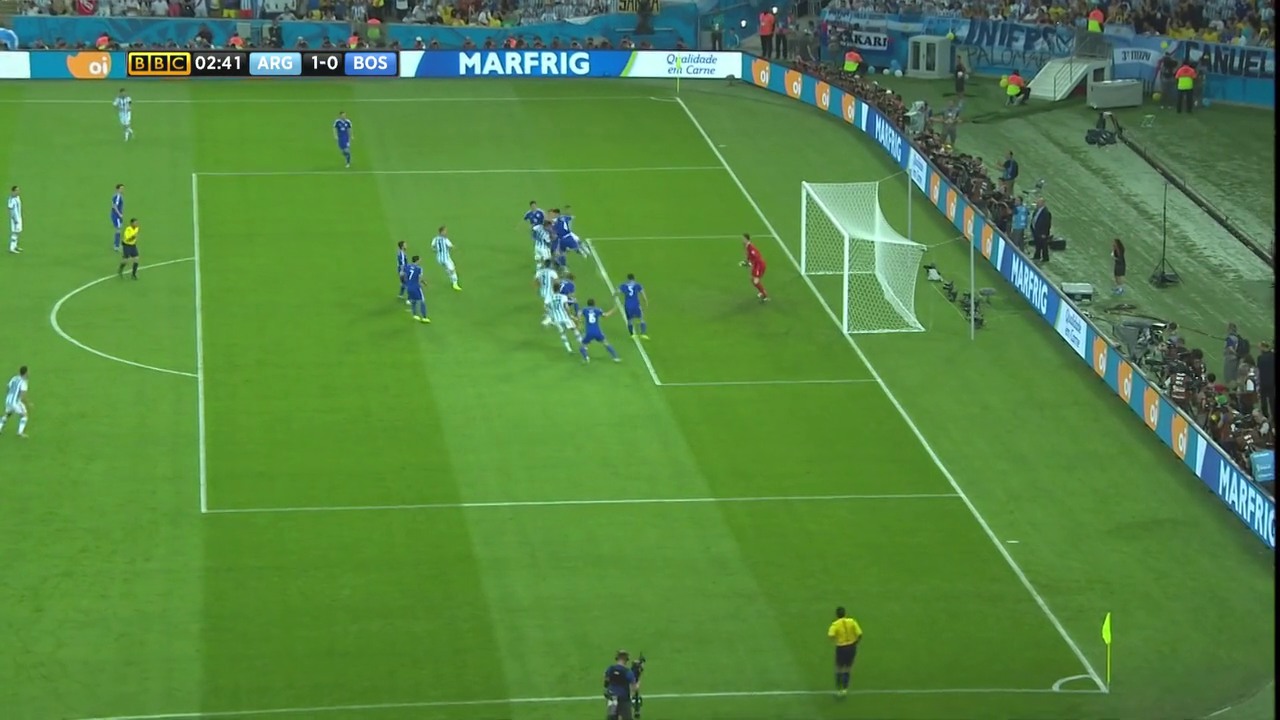} &\hspace{-1.1em}
\includegraphics[width=0.24\linewidth]{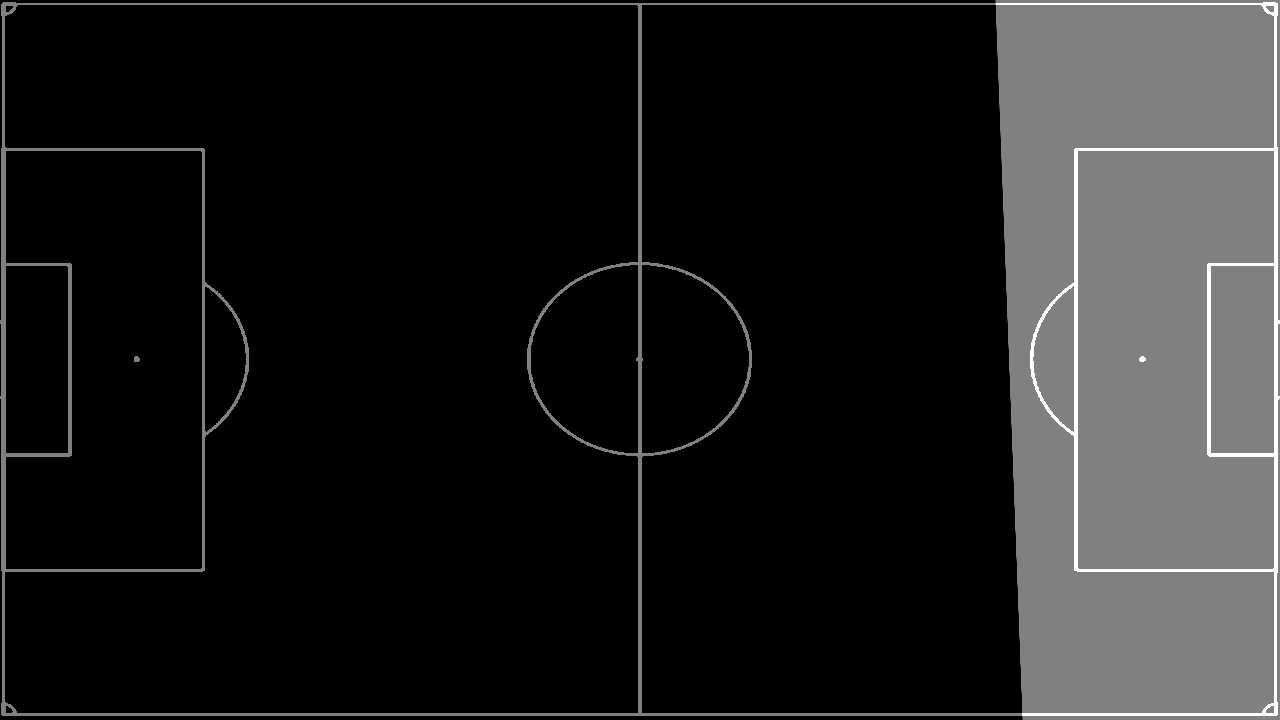} &\hspace{-1em}\vspace{-2.7mm} \\
\multicolumn{2}{c}{\tiny (g)} & \multicolumn{2}{c}{\tiny (h)}\\
\end{tabular}
\caption{Original images and registered static model pairs computed using the HOG based approach. Covering shadows \{(a),(b), (e),(f)\}, motion blur \{(d)\}, varying zoom \{(a),(c)\}, varying camera viewpoints \{(g),(h)\}, varying positions \{(e),(f)\} etc.  }
\label{fig:qualitative_results}
\end{figure}
We selected 500 RGB images from the set of top zoom-out images predicted by the pre-processing algorithm and manually labelled four point correspondences to register it with the static model for quantitative evaluation. The images were selected from 16 different matches. They include varying lighting conditions with prominent shadows, motion blur, varying angles and zooms covering different areas of the playing field to properly test the robustness of the proposed approach. We then evaluate the three approaches over these images and compare them with the corresponding ground truth projections. The IOU measure on the estimated and ground truth projections are given in Table~\ref{table:simulated_results}.

We observe that the HOG features give the best results over the three approaches with a mean IOU measure of around $86$\% (with 91\% of images having IOU measure greater than 75\%). The results degrade by about 5\% from the synthetic case, which occurs due to the limitations of the pre-processing stage to precisely isolate the field lines and remove players and the crowd. The chamfer matching approach seems to be slightly more sensitive to noise. Interestingly, the CNN based approaches degrade considerably over the synthetic experiments. We can draw two conclusions, first that the features from pre-trained networks are succeptible to noise and do not transfer well for the given task. Second, a network trained using synthetic images (where it is easier to create a large training set) may not perform well in presence of noise. To obtain better results in a CNN, we need to train it with a large manually labelled dataset which would help account for the artefacts of the pre-processing stage. On the other hand, HOG with k-NN gives competetive results without this effort. 




%
%
%
\subsubsection{Qualitative evaluation}
\begin{figure*}[t]
\centering
\includegraphics[width=0.95\linewidth]{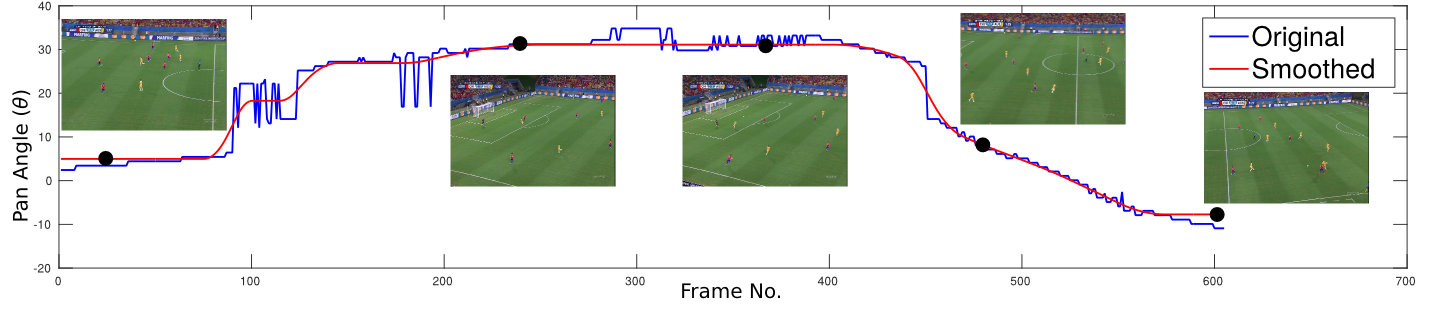}
\caption{Illustration of stabilization using convex optimization. The blue curve shows the pan angle predicted by the proposed approach on each frame individually. The red curve shows the stabilized pan angle after the convex optimization. We can observe the the smoothed pan angle composes of distinct static, linear and quadratic segments. The black dots denote the frames at respective locations.}
\label{fig:convex}
\end{figure*}
Results over a small set of images using HOG based approach are shown in Figure~\ref{fig:qualitative_results}. We can observe that the predictions are quite accurate over diverse scenarios and the method works perfectly even in cases where manual annotation of point correspondences is challenging in itself (Figure~\ref{fig:qualitative_results}(d)). The robustness of our approach over extreme variations in camera angle (Figure~\ref{fig:qualitative_results} (g) and (h)) and challenges like shadows (Figure~\ref{fig:qualitative_results} (a),(b),(e),(f)) and motion blur (Figure~\ref{fig:qualitative_results} (d)) can be observed. The applicability over varying zoom and field coverage is also evident. The reader can refer to the supplementary material for more details, where we provide the results over the entire dataset. 
%
%
%
%
%
%
\subsection{Results over broadcast videos}  
Existing~\cite{okuma2004automatic,lu2013learning} methods for registration are inapplicable for individual frames as they require user to initialize the homography on the first frame with respect to the model, which is then propogated by tracking points in subsequent frames. This requires manual re-initialization of homography at every shot change, in a typical football video of 45 minutes this happens about 400 times. Clearly our method is superior because it is fully automatic. However, we still perform quantitative comparison with the approach in~\cite{okuma2004automatic,lu2013learning} using two long video sequences by manually labelling 200 frames in them (labelling two frames per second) to compute the mean IOU measure. An approach similar to~\cite{okuma2004automatic} which has originally been applied on hockey videos based on KLT tracker gave a IOU measure of 35\% of the football sequences. This low performance can be attributed to drift and lack of features (on the football field as compared to the hockey rink), due to which the tracking fails after few frames. We implemented a more robust variant using SIFT features instead, which gives a mean IOU measure of 70\%. On the other hand, our approach gave a mean IOU measure of 85\% when the registration is computed individually on each frame. 

\paragraph{MRF evaluation:} Using the above mentioned sequences, we then perform MRF optimization by computing k nearest neigbours estimated on the individual frames. We chose k=5 in our experiments.  We found that the  mean IOU measure improved from 85\% to 87\% by employing the MRF based optimization over the per frame results. 

%
%
\paragraph{Convex optimization evaluation:} Qualitative results of the camera stabilization are shown in Figure~\ref{fig:convex} over a video sequence from Chile vs Australia world cup match. The video starts at midfield, pans to left goal post, stays static for few frames and quickly pans back to midfield following a goalkeeper kick. The figure shows the pan angle trajectory of the per frame predictions with and without camera stabilization. We can observe that the optimization clearly removes jitter and replicates a professional cameraman behaviour. The actual and the stabilized video are provided in the supplementary material.
%

%
%
%
\section{Summary}
We have presented a method to compute projective transformation between a static model and a broadcast image as a nearest neighbour search and have shown that the presented approach gives highly accurate results (about 87\% after MRF smoothing) over challenging datasets. Our method is devoid of any manual initialization prevalent in previous approaches ~\cite{okuma2004automatic,lu2013learning}. Once the dictionary is learnt, our method can be directly applied to any standard football broadcast and in fact can be easily extended to any sport where such field lines are available (like basketball, ice hockey etc.). Moreover, the semi supervised dictionary generation allows us to adapt the algorithm even if new camera angles are used in future. The proposed method opens up a window for variety of applications which could be realized using the projected data. One limitation of our approach is that it is only applicable to top zoom-out views and it would be an interesting problem to register other kind of shots (ground zoom-in, top zoom-in shots) using the predictions over top zoom out views, player tracks and other temporal cues. 
{\small
\bibliographystyle{ieee}
\bibliography{eccv2016_soccer}
}

\end{document}